\documentclass[10pt,twocolumn,letterpaper]{article}

\usepackage{iccv}
\usepackage{times}
\usepackage{epsfig}
\usepackage{graphicx}
\usepackage{amsmath}
\usepackage{amssymb}
\usepackage{color}

\usepackage{multirow}
\usepackage{diagbox} 
\usepackage{enumitem}

\newcommand{\minisection}[1]{\vspace{0.04in} \noindent {\bf #1}}
\newcommand*{\affmark}[1][*]{\textsuperscript{#1}}

\usepackage[pagebackref=true,breaklinks=true,letterpaper=true,colorlinks,bookmarks=false]{hyperref}

\iccvfinalcopy % *** Uncomment this line for the final submission

\ificcvfinal\fi

\begin{document}

%%%%%%%%% TITLE
\title{TransferI2I: Transfer Learning for Image-to-Image Translation from Small Datasets}
% with limited data ?

\author{Yaxing Wang\affmark[1], Héctor Laria Mantecón\affmark[1]\\
Joost van de Weijer\affmark[1], Laura Lopez-Fuentes\affmark[2], Bogdan Raducanu\affmark[1]\\
\affmark[1]Computer Vision Center, Universitat Aut\`onoma de Barcelona, Spain \\
\affmark[2]Universitat de les Illes Balears, Spain \\
{\tt\small \{yaxing,hlaria,joost,bogdan\}@cvc.uab.es, l.lopez@uib.es}
% \and
}

\maketitle
\ificcvfinal\thispagestyle{empty}\fi

%%%%%%%%% ABSTRACT
\begin{abstract}
  Image-to-image (I2I) translation has matured in recent years and is able to generate high-quality realistic images.  However, despite current success, it still faces important challenges when applied to small domains.  Existing methods use transfer learning for I2I translation, but they  still require  the learning of millions of  parameters from scratch. This drawback severely limits its application on small domains.
In this paper, we propose a new transfer learning for I2I translation (TransferI2I). We decouple our learning process into the image generation step and the I2I translation step.  In the first step we propose two novel techniques:  source-target initialization and self-initialization of the adaptor layer. The former finetunes the pretrained generative model (e.g., StyleGAN) on source and target data. The latter  allows to initialize all non-pretrained network parameters without the need of any data. These techniques provide a better initialization for the I2I translation step. 
In addition, we introduce an auxiliary GAN that further facilitates the training of deep I2I systems even from small datasets. In extensive experiments on three datasets, (Animal faces, Birds, and Foods), we show that we outperform existing methods and that mFID improves on several datasets with over 25 points.  Our code is available at: \url{https://github.com/yaxingwang/TransferI2I}. 
\end{abstract}

%%%%%%%%% BODY TEXT
\section{Introduction}\label{sec:introduction}
Image-to-image (I2I) translation aims to map an image from a source to a target domain. Several methods obtain outstanding results on paired data~\cite{pix2pix2017,zhu2017toward}, unpaired data~\cite{kim2017learning,yi2017dualgan,zhu2017unpaired}, scalable I2I translation~\cite{choi2020stargan,liu2018unified,perarnau2016invertible}  and diverse I2I translation~\cite{choi2020stargan, huang2018multimodal,Lee2018drit}. \textit{Scalable I2I} translation aims to translate images between multiple domains. For example, a cat face is mapped onto other animal faces (i.e. dog, tiger, bear, etc.). The goal of \textit{diverse I2I} translation is to synthesize multiple plausible outputs of the target domain from a single input image (i.e. translating a dog face to various plausible cat faces). Despite impressive leaps forward with paired, unpaired, scalable and diverse I2I translation, there are still important challenges. Specifically, to obtain good results existing works rely on large labelled data. When given small datasets (e.g., 10 images per domain) current algorithms suffer from inferior performance. Also, labeling large-scale datasets is costly and time-consuming, making those methods less applicable in practice. 

Several works~\cite{benaim2021structural,benaim2017one,cohen2019bidirectional,lin2020tuigan,liu2019few} have studied  one-shot and few-shot I2I translation.  One-shot I2I translation~\cite{benaim2021structural,benaim2017one,cohen2019bidirectional,lin2020tuigan} refers to the case where  only \textit{one source} image and \textit{one or few}  target images are available. These works fail to perform  multiclass I2I  translation. FUNIT~\cite{liu2019few} conducts few-shot I2I translation, but still requires large datasets at the training stage. In this paper, we focus on transfer learning for I2I translation with limited data.

Recent work~\cite{shocher2020semantic,wang2020deepi2i} leverages transfer learning for I2I translation. SGP~\cite{shocher2020semantic} utilizes a pretrained classifier (e.g, VGG~\cite{simonyan2013deep}) to initialize the encoder of an I2I model. However, the remaining networks (i.e., decoder, discriminator and adaptor layers\footnote{We follow~\cite{wang2020deepi2i} and call the layers which connect encoder and decoder at several levels adaptor layers.}) need to be trained from scratch, which still requires a large dataset to train the I2I translation model. DeepI2I~\cite{wang2020deepi2i} uses a pretrained GAN (e.g., StyleGAN~\cite{karras2019style} and BigGAN~\cite{brock2018large}) to initialize the I2I  model. However, it still requires to train the adaptor layers from scratch. The adaptor layers contains over 85M parameters (using the pretrained BigGAN) which makes their training on translation between small domains prone to overfitting. Since both SGP and DeepI2I leverage the adaptor between the encoder and the generator, one potential problem is that the generator easily uses the information from the high-resolution skip connections (connecting to the upper layers of the generator), and ignore the deep layers of the generator, which require a more semantic understanding of the data, thus more difficult to train. 
Inspired by DeepI2I, we use the pretrained GANs to initialize I2I translation model. Differently, we propose a new method to train I2I translation, overcoming the overfitting and improving the training of  I2I model.

\begin{figure*}[t]
    \centering
    \includegraphics[width=0.8\textwidth]{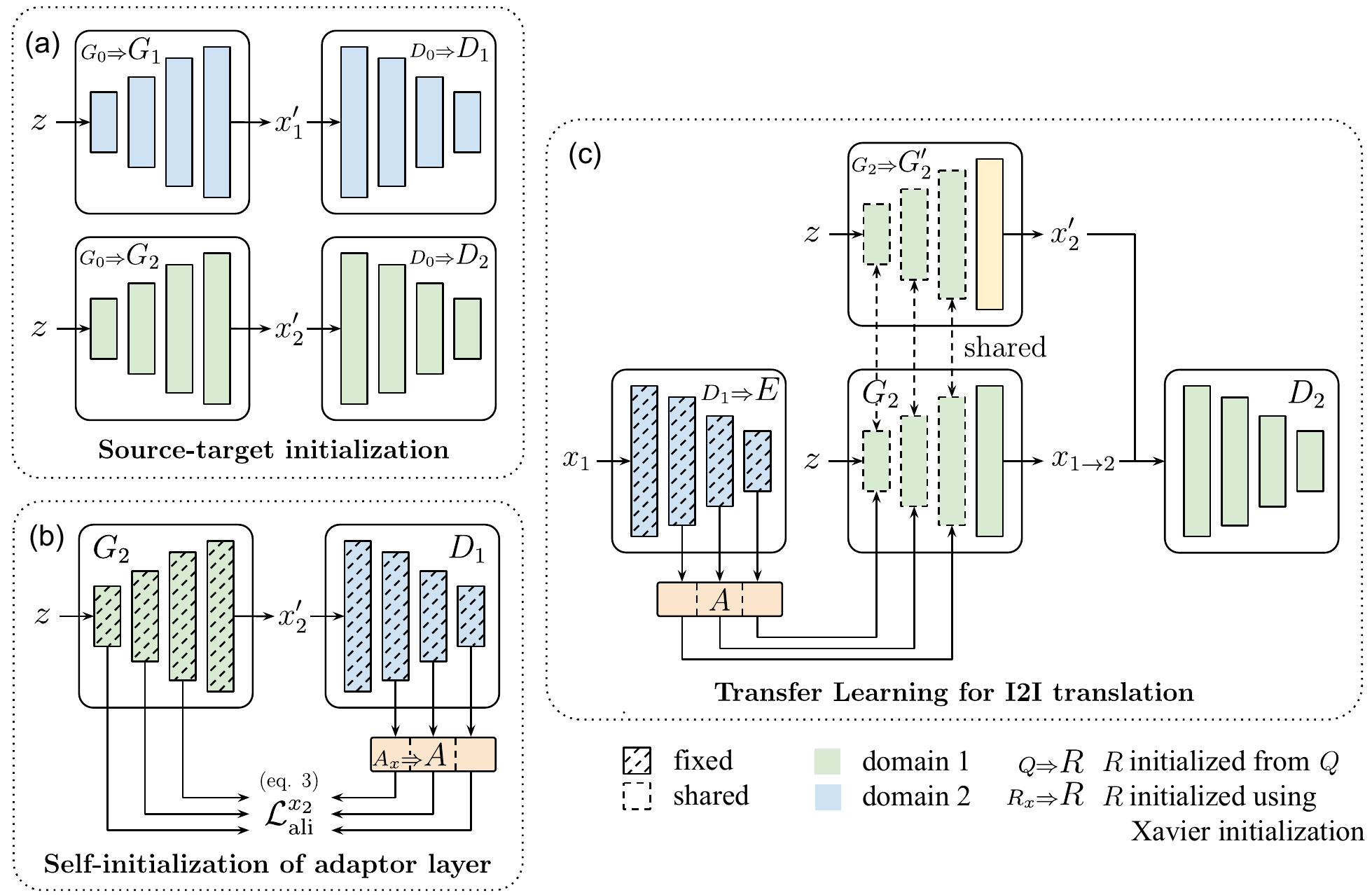}
    \caption{\small    \small Model architecture and training stages. Here modules come from the immediate previous stage unless otherwise indicated. A pretrained GAN (e.g., StyleGAN~\cite{karras2019style}) is used as $G_0$ and $D_0$ to initialize the two GANs. (a) \textit{Source-target initialization} performs finetuning on two domains (i.e., $\mathcal{X}_1$ and  $\mathcal{X}_2$) to form two independent GANs (i.e., the generator $G_1$ and the discriminator $D_1$, the generator $G_2$ and the discriminator $D_2$). (b) \textit{Self-initialization} of adaptor layer to pretrain the adaptor $A$ and align both the generator $G_2$ and the discriminator $D_1$.   We only update the adaptor layers $A$. (c) The I2I translation model is composed of five main parts: the encoder $E$, the adaptor layer $A$, the generator $G_2$, the \textit{auxiliary generator} $G_2'$ and the  discriminator $D_2$. Note the encoder $E$ is initialized by the discriminator $D_1$. The portion of weights from $G'_2$ that is not shared (in yellow), is initialized with $G_2$ weights.
    }\vspace{-3mm}
    \label{fig:models}
\end{figure*}

In this paper, we  decouple our learning process into two steps:  image generation and I2I translation. The first step aims to train a better generative model, which is leveraged to initialize the I2I translation system, and contributes to improve I2I translation performance. We introduce two contributions to improve the efficiency of the transfer, especially important for small domains. (1) we improve \textit{source-target initialization} by finetuning the pretrained generative model
(e.g., StyleGAN) on source and target data.
This ensures that networks are already better prepared for their intended task in the I2I system. (2) we propose a \emph{self-initialization} to pretrain the weights of the adaptor networks (the module $A$ in Figure~\ref{fig:models} (b)) without the need of any data. Here, we exploit the fact that these parameters can be learned by generating the layer activations from both the generator and discriminator (by sampling from the latent variable $z$). From these activations the adaptor network weights can be learned. For the second step we conduct the actual I2I translation using the learned weights in the first step. Furthermore, we propose an \textit{auxiliary generator} to  encourage the usage of the deep layers of the I2I network. 

Extensive experiments on a large variety of datasets confirms the superiority of the proposed transfer learning technique for I2I. It also shows that we can now obtain high-quality image on relatively small domains. This paper shows that transfer learning can reduce the need of data considerably; as such this paper opens up application of I2I to domains that suffer from data scarcity. Our main contributions are: 
\setlist{nolistsep}
\begin{itemize}[noitemsep]
      \item We explore I2I translation with limited data, reducing the amount of required labeled data.
       \item We propose several novel techniques (i.e., \textit{source-target initialization}, \textit{self-initialization} and \textit{auxiliary generator}) to facilitate this challenging setting.
        \item We extensively study the properties of the proposed approaches on two-class and multi-class I2I translation tasks and achieve significant performance improvements even for high quality images.  
\end{itemize}

\section{Related work}\label{sec:related_work}
\minisection{Generative adversarial networks.}  GANs~\cite{goodfellow2014generative} are a combination of a generator $G$ and a discriminator $D$. The goal of the generator is to learn a mapping from a latent code, i.e. a noise source, to the training data distribution. 
Conversely, the discriminator network, or critic \cite{arjovsky2017wasserstein}, learns to distinguish between the real data and generated instances from $G$ in the fashion of an adaptive loss. In this opposed game, both networks improve upon each other to the point of yielding state-of-the-art image generation. Recent works~\cite{arjovsky2017wasserstein,gulrajani2017improved,mao2017least} aim to overcome mode collapse and training instability %, which occurs %when 
problems, which frequently occur when optimizing GANs. Besides, several works~\cite{brock2018large,denton2015deep,karras2019style} explore constructing effective architectures to synthesize high-quality images. 

\minisection{I2I translation.}
Image-to-image translation has been widely studied in computer vision. It has achieved outstanding performance on  both paired~\cite{gonzalez2018image,isola2016image,zhu2017toward} and unpaired image translation~\cite{kim2017learning,liu2017unsupervised,mejjati2018unsupervised,park2020contrastive,yi2017dualgan,zhu2017unpaired}. These approaches, however, face two main challenges: diversity and scalability.  The former aims to generate  multiple plausible outputs of the target domain from a single input image~\cite{almahairi2018augmented,huang2018multimodal,Lee2018drit,yu2019multi}. The goal of scalable I2I translation is to map images~\cite{choi2020stargan,lee2020drit++,yu2019multi} across several domains using a single model. 
Several works~\cite{katzir2019cross,saito2020coco,shocher2020semantic,wu2019transgaga} explore the difficult task: the \textit{shape} translation as well as \textit{style}.
TransGaGa~\cite{wu2019transgaga} disentangles the source image into two spaces:  the geometry and style. Then it conducts the translation for each latent space separately.  However, none of these approaches addresses the problem of transfer learning for I2I.

Several recent works used GANs for I2I translation with few test samples. 
Lin \etal proposed a zero-shot I2I translation method, which leverages pairs of images and captions to study domain-specific and domain-invariant features. Recent work~\cite{benaim2021structural,benaim2017one,cohen2019bidirectional,lin2020tuigan} explore one-shot I2I translation, and propose one-shot specific I2I models. However, these methods cannot be used for multi-class I2I translation, since these model are designed for  the two-class case where a few images of two domains can be accessed.  FUNIT~\cite{liu2019few} is the first to study few-shot I2I translation, but still relies on vast quantities of labeled source domain images  for training.  % from source to target domain.  

\minisection{Transfer learning.}
Transfer learning aims to reduce the training time, improve the performance and reduce the amount of training data required by a model by reusing the knowledge from another model which has been trained on another, but related, task or domain.
A series of recent works investigated knowledge transfer on generative models~\cite{noguchi2019image,wang2018transferring} as well as discriminative models~\cite{donahue2014decaf}. More recent work~\cite{shocher2020semantic,wang2020deepi2i} explore the knowledge transfer for I2I translation. Both methods, however, introduce a new network module  which is trained from scratch, and prone to suffer from overfitting. Several other approaches~\cite{goetschalckx2019ganalyze,jahanian2019steerability}  perform the image manipulation based on the pretrained GAN. Especially, given the pretrained generator (e.g., StyleGAN) they expect to  manipulate  the output image attribute (e.g., the age, the hair style, the face pose etc..). However, these methods do not focus on transfer learning.  
Furthermore, some methods~\cite{abdal2019image2stylegan,bau2018gan,zhu2020domain} embed a given exemplar image into the input latent space of the pretrained GAN (e.g., StyleGAN). These methods literately  optimize the latent space to reconstruct the provided image. In fact, they do not perform I2I translation.  

\section{Method}\label{sec:method}

\minisection{Problem setting.} Our goal is to propose an unpaired I2I translation system for the case when training data is limited.
Therefore, we apply transfer learning to improve the efficiency of I2I translation. We decouple our learning into an image generation step and a I2I translation step. The image generation step contains two contributions: (a) \emph{source-target initialization}, and (b) adaptor layer \textit{self-initialization}. In addition, for the I2I step, we introduce an \textit{auxiliary generator} to address the inefficient usage of the deep layers of the generator when using the skip connections in I2I model.

Figure~\ref{fig:models} provides an overview of our method. Figure~\ref{fig:models}(a) shows the  \emph{source-target initialization}, in which we learn a better generation model, thus contributing to a better initialization of the I2I translation models. Next, in Figure~\ref{fig:models}(b), we introduce \textit{self-initialization} to overcome overfitting of the adaptor layers. In Figure\ref{fig:models}(c), we train the I2I network, and introduce an \textit{auxiliary generator}. This additional generator shares several layers with the main generator, encouraging the usage of the under-performing deep layers.  Our method is general for two-class I2I translation and multi-class I2I translation.  First we introduce our method for single-class I2I (Section~\ref{sec:two_class_overview}) and in the next section we extend our method to multi-class I2I (Section~\ref{sec:many_class_overview}).

\subsection{Method overview}\label{sec:two_class_overview}
We consider two domains: source domain $\mathcal{X}_1 \subset \mathbb{R}^{H\times W\times 3} $ and target domain $\mathcal{X}_2 \subset \mathbb{R}^{H\times W\times 3} $. In this work, given limited training samples from both source and target domains, we aim to map a source image  $x_1  \in \mathcal{X}_1 $ into a target sample $x_{1 \rightarrow 2} \in \mathcal{X}_2$.   Let   vector $\mathbf{z \in \mathbb{R}^{Z}}$ be random noise. 

\minisection{Source-target initialization.} Given a pretrained GAN  (e.g., StyleGAN) and limited training data,
we propose to train two generative models on the limited available data of respectively the source and target domain (Figure~\ref{fig:models}~(a)).
Especially,  we train a generative model which is composed of a generator $G_i$ and a discriminator $D_i$ for each domain $\mathcal{X}_i ,i = \{1, 2 \} $. 
Here we apply finetuning to adapt to the source and target domains like in ~\cite{wang2018transferring}. This step could be further improved by using recent approaches to improve the transfer on small domains~\cite{noguchi2019image,mo2020freeze,zhao2020leveraging,li2020few}.
The training objective becomes:

\begin{equation}
\begin{aligned}
\mathcal{L}_{GAN}^{x_1} &= \mathbb{E}_{x_1\sim \mathcal{X}_1}\left[ \log D_1 \left ( \mathbf{x_1} \right ) \right]  \\
&+  \mathbb{E}_{\mathbf{z} \sim p(\mathbf{z})}\left[ \log \left ( 1 - D_1 \left ( G_1 \left ( \mathbf{z} \right ) \right ) \right )   \right]
\end{aligned}
\end{equation} 
\begin{equation}
\begin{aligned}
\mathcal{L}_{GAN}^{x_2} &= \mathbb{E}_{x_2\sim \mathcal{X}_2}\left[ \log D_2 \left ( \mathbf{x_2} \right ) \right]  \\
&+  \mathbb{E}_{\mathbf{z} \sim p(\mathbf{z})}\left[ \log \left ( 1 - D_2 \left ( G_2 \left ( \mathbf{z} \right ) \right ) \right )   \right] .
\end{aligned}
\end{equation} 
Here the generative models for both the source and target domains are used to provide a better initialization for the I2I translation.  

\minisection{Self-initialization of adaptor layer.}\label{sec:alignment} Inspired by DeepI2I~\cite{wang2020deepi2i}, we use the pretrained discriminator (Figure~\ref{fig:models}(a)) to initialize both the encoder and the discriminator of the I2I model (Figure~\ref{fig:models}(c)), and correspondingly, the pretrained generator to the I2I generator.
Since in the GAN configuration there are no connections between intermediate layers of the generator and discriminator, these layers are not aligned. For that reason,~\cite{wang2020deepi2i} introduces an adaptor network (indicated by $A$ in Figure~\ref{fig:models}(b,c)) to communicate between the various layers of the encoder and decoder.
In DeepI2I, they found that the introduction of four adaptor networks is optimal. These layers contain a significant amount of the total number of parameters in the system (around 25\%). They then proceed to naively optimize these parameters on the source-target data. Training the adaptor network from scratch leads to overfitting on limited data. 

To overcome these drawbacks, we propose a procedure,  called \emph{self-initialization}, that leverages the previous pretrained model (Figure~\ref{fig:models}(a)) to align the adaptor networks without the need of any data.  As shown in Figure~\ref{fig:models}(b),  the noise $\mathbf{z}$ is taken as input for the generator $G_2$, from which we extract the hierarchical representation $F_{g}(\mathbf{z})  = \left \{ G_2(\mathbf{z})_{l} \right \}$ as well as  the synthesized image $G_2(\mathbf{z})$. Here $G_2(z)_{l}$ is the $l_{th} (l = m,\ldots,n , (n>m))$ ResBlock~\footnote{After each ResBlock the feature resolution is half of the previous one in both encoder and discriminator,  and  two times in generator} output of the generator $G_2$. We then take the generated image $G_2(\mathbf{z})$ as input for the discriminator $D_1$, and similarly  collect the hierarchical feature  $F_{d}(\mathbf{z}) = \left \{ D_1(G_2(\mathbf{z}))_{l} \right \}$. The adaptor network $A$ finally takes the output representation  $\left \{ D_1(G(\mathbf{z}))_{l} \right \}$ as input, that is $A(F_{d}(\mathbf{z})) = \left \{ A \right \} $. In this step, our loss is:
\vspace{-3pt}
\begin{equation}
\begin{aligned}
 \mathcal{L}_{ali}^{x_2}= \sum _{l}  \left \| F_g \left ( z \right ) - A \left ( D_1(G_2(z)) \right )  \right \|_{1}.
\end{aligned}
\end{equation}
In this step both the generator and the discriminator are frozen and only the adaptor layers are learned. Note that the adaptor layers are trained to take the discriminator as input, and output a representation that is aligned with the generator (opposite to the order in which they are applied in the GAN); this is done because the generator and discriminator are switched in the I2I network (see Figure~\ref{fig:models}(c)) when we use the pretrained discriminator to initialize the encoder.

\minisection{Transfer Learning for I2I translation.} Figure~\ref{fig:models}(c) shows how to map the image from the source domain to target domain. For example, to translate a source image $x_1  \in \mathcal{X}_1 $ to $x_{1 \rightarrow 2} \in \mathcal{X}_2$.  Our architecture consists of 5 modules: the encoder $E$, the adaptor $A$, the generator $G_{2}$, the \textit{auxiliary generator} $G_{2}'$ and the discriminator $D_{2}$. Let $E_{l}$ be the $l_{th} (l = m,\ldots,n , (n>m))$ ResBlock output of the encoder $E$, which is further taken  as input for the corresponding adaptor network $A_{l}$. 

We aim to map the image  from the source to the target domain with limited labeled data. First, the encoder $E$,  initialized by the pretrained discriminator $D_1$ takes the  image $x_1$ as input, extracting the hierarchical representation  $E_{g}(x_1) = \left \{ E(x_1)_{l} \right \}$ from different layers,  which contains both the structural and semantic information of the input image. 
$E_{g}(x_1)$ is then fed to the adaptor network $A(x_1) = \left \{ A(x_1)_{l} \right \}$, which in turn is taken as input  for the generator $G_2$ along with the noise $z$ to synthesize the output image $x_{1 \rightarrow 2} = G_{2}(z, A(E(x_1)))$ (for details see Supplementary Material~\ref{supp:adaptor}).  We employ the discriminator $D_2$ to distinguish real images from generated images, and preserve a similar pose in input source image $x_1$ and the output $G_2(z, A(E(x_1)))$~\cite{liu2019few,wang2020deepi2i}. 

Training the I2I translation model can lead to unused capacity of the deep layers of the generator, largely due to the skip connections.  It is relatively easy for the generator to use the information from the high-resolution skip connections (connecting to the upper layers of the generator), and ignore the deep layers of the generator, which require a more semantic understanding of the data, thus more difficult to train. 
To address this, we propose an \emph{auxiliary generator} which has the same network design, but only uses the noise as input. Taking the translation from the source image $x_1  \in \mathcal{X}_1 $ to $x_{1 \rightarrow 2} \in \mathcal{X}_2$ as example. The \textit{auxiliary generator} $G_2'$ takes the noise $z$ as input, and synthesizes the output image $x_2' \in \mathcal{X}_2 $. We propose to share the deep layers of this \textit{auxiliary generator} with the ones following the skip connection in the main generator $G_2$ (the dashed layers in Figure~\ref{fig:models}(c)). Since $G_2'$ has no access to skip connections, it is forced to use its deep layers, and since we share these, the main I2I generator is also driven to use them.

Our loss function for I2I translation is a multi-task objective comprising: (a) \textit{adversarial loss} which classifies the real image and the generated image. (b) \textit{reconstruction loss}  guarantees that both the input image $x_1$ and the synthesized image $x_{1 \rightarrow 2} = G_{2}(z, A(E(x_1)))$ keep the similar structural information.

\minisection{\textit{Adversarial loss.}} We employ GAN~\cite{goodfellow2014generative}  to optimize this problem as follows:
\begin{equation}
\begin{aligned}
\mathcal{L}_{GAN}^{x_2} &= \mathbb{E}_{x_2\sim \mathcal{X}_2}\left[ \log D_2 \left ( \mathbf{x_2} \right ) \right]  \\
&+   \mathbb{E}_{\mathbf{x_1} \sim \mathcal{X}_1, \mathbf{z} \sim p(\mathbf{z})}\left[ \log (1 - D_2 \left ( G_2 \left ( A \left ( E \left ( \mathbf{x_1} \right ) \right ), \mathbf{z} \right )  \right ) \right]  \\
&+ \lambda_{aux} \mathbb{E}_{\mathbf{z} \sim p(\mathbf{z})}\left[ \log \left ( 1 - D_2 \left ( G_2' \left ( \mathbf{z} \right ) \right ) \right )   \right],
\end{aligned}
\end{equation} 
where $p \left ( \mathbf{z} \right )$ follows the normal distribution. The hyper-parameter $\lambda_{aux}$ is to balance the importance of each terms. We set $\lambda_{aux} = 0.01$.  The discriminator $D_1$ and loss $\mathcal{L}_{GAN}^{x_1}$ are similar.

\minisection{\textit{Reconstruction loss.} } We use reconstruction  to preserve the structure of both the input image $x_1$  and the output image $x_{1 \rightarrow 2}$. In the same fashion  as results for photo-realistic image generation~\cite{jakab2018unsupervised,johnson2018image,shrivastava2017learning}, we  use the discriminator output to achieve this goal through the following loss:

\begin{equation}
\begin{aligned}
 \mathcal{L}_{rec}^{x_1} &= \sum _{l} \alpha_{l} \left \lVert D_2 \left ( \mathbf{x_1} \right ) - D_2 \left ( \mathbf{x_{1 \rightarrow 2}} \right )  \right \rVert_{1} ,
\end{aligned}
\end{equation}
where parameters $\alpha_{l}$ are scalars which balance the terms. Note we set $\alpha_{l} = 1$.

\minisection{\textit{Full Objective.}}
The full objective function of our model is:
 \vspace{-2pt}
\begin{equation}
\label{eq:loss}
\begin{aligned} 
&\min_{\substack{E_{1},E_{2}, A_{1}, A_{2}\\ G_{1}, G_{1}^{'}, G_{2}, G_{2}^{'}}} \underset{D_1, D_2}{\max}   \mathcal{L}_{GAN}^{x_1} + \mathcal{L}_{GAN}^{x_2}+ \lambda_{rec} \left ( \mathcal{L}_{rec}^{x_1}  + \mathcal{L}_{rec}^{x_2} \right )
\end{aligned}
\end{equation}
where $\lambda_{rec}$ is hyper-parameters that balance the importance of each terms.  We set $\lambda_{rec} = 1$.

\subsection{Multi-class I2I translation}\label{sec:many_class_overview}
Our method can also be applied to multi-class I2I translation. Comparing with the \emph{source-target initialization}  of the two-class I2I translation (Figure~\ref{fig:models} (a)), we instead have one conditional generator and one conditional discriminator by using a class embedding like BigGAN~\cite{brock2018large}. Especially, we perform  \emph{source-target initialization} from the pretrained conditional GAN (e.g., BigGAN), and obtain a single generator and discriminator for all data. The following steps, the \textit{self-initialization of the adaptor} (Figure~\ref{fig:models}(b)) and the I2I translation with the \textit{auxiliary generator} (Figure~\ref{fig:models}(c)), are similar to the ones of two-class I2I system except for the conditioning for both generator and discriminator. The framework for multi-class I2I translation is shown in Supp.~Mat.~A.

\begin{table}[t]

    \setlength{\tabcolsep}{1mm}
    \resizebox{\columnwidth}{!}{%
    \centering
    \footnotesize
    \setlength{\tabcolsep}{5pt}
    \begin{tabular}{|c|c|c|c|c|c|c|c|c|}
 \hline
         &Network & Optimizer	& Lr		&$(\beta_{1},\beta_{1}$)&Bs	&Is \cr\cline{1-7}
      \multirow{2}{*}{Two-class I2I} &$G$&Adam&$1e^{-5}$&(0.0,0.99)&16&256\cr\cline{2-7}  
     &$A,D$&Adam&$1e^{-3}$&(0.0,0.99)&16&256\cr\cline{2-7}  
   \hline
      \multirow{2}{*}{Multi-class I2I} &$G$&Adam&$5e^{-5}$&(0.0,0.999)&16&128\cr\cline{2-7}  
     &$A,D$&Adam&$2e^{-4}$&(0.0,0.999)&16&128\cr\cline{2-7}  
   \hline
    \end{tabular}  
    }
\caption{\small   The experiment configuration. Lr: learning rate, Bs: batch size, Is: image size.}\vspace{-6mm}\label{tab:network_configure}
\end{table}

\section{Experiments}\label{sec:experiment}
In this section, we first introduce the experimental settings (Section~\ref{sec:experiment_setting}): the training details, evaluation measures, datasets and baselines. We then evaluate our method on two cases:  \textit{multi-class I2I translation} (Section~\ref{sec:many_class_case}) and \textit{two-class I2I translation} (Section~\ref{sec:two_class_case}). 

\subsection{Experiment setting}\label{sec:experiment_setting}

\minisection{Training details.}
We adapt the structure of the pretrained GAN (i.e., StyleGAN for two-class I2I translation and BigGAN for multi-class I2I translation) to our architecture. Especially, both the generator $G$ and the discriminator $D$ directly duplicate the ones of the GAN (i.e., StyleGAN or BigGAN). The auxiliary generator $G'$ is same to the generator, and the encoder $E$ copy the structure of the  discriminator. The adaptor network $A$ contains four sub-adaptor networks. In multi-class I2I system, each of the sub-adaptor consists of one Relu, two convolutional layers (Conv) with $3 \times 3$ filter size and stride of 1,  and one Conv with $1 \times 1$ filter and stride of 1, except for the fourth sub-adaptor (corresponding to the deepest layer of the encoder) which only contains two Convs with $3 \times 3$ filter and stride of 1.  In two-class I2I system, each sub-adaptor network is composed of Conv with  $3 \times 3$ filter size and stride of 1.   The proposed method is implemented in Pytorch~\cite{paszke2017automatic}.  The configure of the experiment is reported in Table~\ref{tab:network_configure}. We use $1\times$ Quadro RTX 6000 GPUs (24 GB VRAM to conduct all our experiments.

\begin{table}[t]
    \setlength{\tabcolsep}{1pt}\renewcommand{\arraystretch}{1}    \resizebox{\columnwidth}{!}{%
    \centering
    \footnotesize
    \setlength{\tabcolsep}{5pt}
    \begin{tabular}{|c|c|c|c|c|}
    \hline
    source-target initialization & Self-initialization  & mKID $\times$100$\downarrow$ &  mFID $\downarrow$  \cr
    \hline
    % \multirow{}{*}{}
      $\times$   & $\times$   & 11.48	& 137.11	 \cr\cline{1-4}\hline
      $\surd$   & $\times$   & 9.63	& 114.23	 \cr\cline{1-4} \hline
       $\times$   & $\surd$       & 10.03 & 122.12 \cr\cline{1-4}\hline
        $\surd$   & $\surd$          &9.40	& 109.7 \cr\cline{1-4}\hline
    \end{tabular}   
    }
 \caption{\small Influence of \emph{source-target initialization} and \emph{self-initialization} of the adaptor on \emph{Animal faces}.}\vspace{-7mm}\label{tab:abalation_FT_AD_AU}
\end{table}

\minisection{Evaluation metrics.}
We use several GAN metrics. The first one is Fr\'echet Inception Distance (FID)~\cite{heusel2017gans}, which compares the distributions of the real and fake images using the Fr\'echet distance. The second one is Kernel Inception Distance (KID)~\cite{binkowski2018demystifying}, which calculates the maximum mean discrepancy (MMD) of the same embedded features and is proven to be a converging estimator, contrary to FID. To account for all categories, we calculate the mean FID and KID as \textit{mFID} and \textit{mKID}. %
Finally, we train a real (\textit{RC}) and a fake classifier (\textit{FC})~\cite{shmelkov2018good} to evaluate the ability to generate class-specific images. RC is trained on real data and evaluated on the generated data and vice versa for FC.

\minisection{Datasets.}
We evaluate our method on five datasets. For multi-class I2I translation, we use three datasets: \emph{Animal faces}~\cite{liu2019few}, \emph{Birds}~\cite{van2015building} and \emph{Foods}~\cite{kawano2014automatic}. To evaluate the two-class I2I model, we use two datasets: \emph{cat2dog-200}~\cite{lee2020drit++} and \emph{cat2dog-2035}~\cite{lee2020drit++}.  The \emph{Animal faces} dataset contains 1,490 images and 149 classes in total, \emph{Birds} has 48,527 images and 555 classes in total, \emph{Foods} consists of 31,395 images and 256 classes in total.  We resized all images from the \textit{Animal faces}, \textit{Birds} and \textit{Foods} to $128  \times 128$, and split each data into a training (90 $\%$) and test set (10 $\%$) except for the \textit{Animal faces} in which the number of  test images  is 1,490 (10/per class).  The \emph{cat2dog-200} is composed of 200 images (100 images/per class). The \emph{cat2dog-2035} contains 771 images for the cat category and 1264 images for the dog category. The test dataset for both  \emph{cat2dog-200} and \emph{cat2dog-2035} is the same, and has 200 images (100 images/per class) with an image size of $256 \times 256$.
\begin{figure}[t]
    \centering
    \includegraphics[width=\columnwidth]{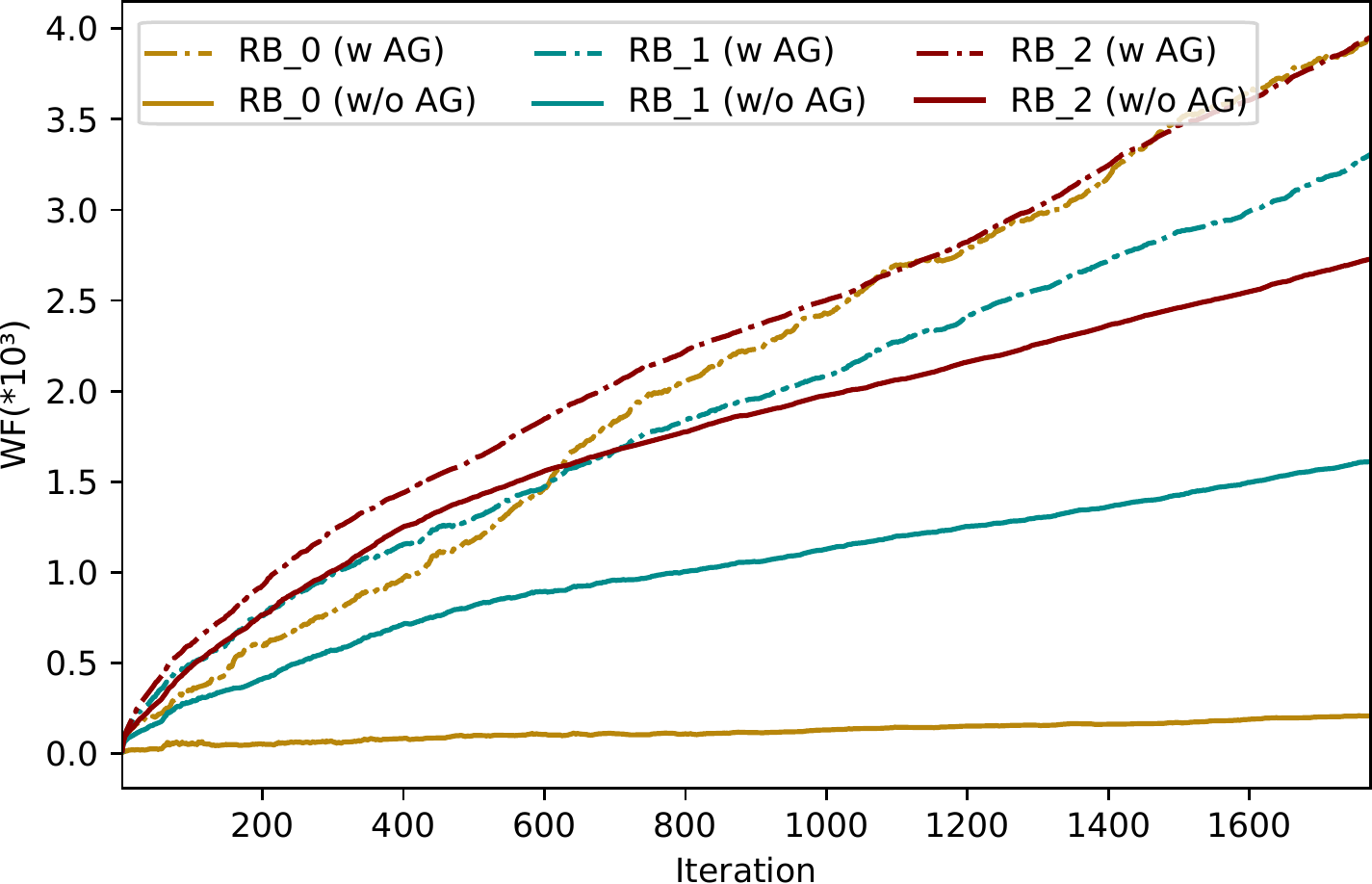} 
    \caption{\small The change of the weights w (w/o) the \textit{auxiliary generator} (AG). \textit{RB\_i} ($i=0,1,2$) is the index  of ResBlock layer of the generator from input to the output. \textit{WF} is the weight fluctuation.}\vspace{-3mm}
    \label{fig:ewc}
\end{figure}

\minisection{The baselines for two-class I2I.}
We compare to several baselines for this setting. \textit{CycleGAN~\cite{zhu2017unpaired}} first perform unpaired I2I translation  by leveraging a cycle consistency loss to reconstruct the input image.
\textit{UNIT~\cite{liu2017unsupervised}} presents an unsupervised I2I translation method under the shared-latent space assumption.
The related methods, including \textit{MUNIT~\cite{huang2018multimodal}},  \textit{DRIT++}~\cite{lee2020drit++} and \textit{StarGANv2}~\cite{choi2020stargan}, propose disentanglement to control separately the pose and style  information. 
 \textit{NICEGAN~\cite{chen2020reusing}} proposes a sharing mechanism between the encoder and the discriminator. \textit{UGATIT~\cite{kim2019u}}  aims to handle the geometric changes, and introduce two techniques: an attention module and a new normalization. CUT~\cite{park2020contrastive} introduces contrastive learning  for I2I translation. DeepI2I~\cite{wang2020deepi2i} uses pretrained GANs to initialize the I2I model.

\begin{table}[t]
    \setlength{\tabcolsep}{1pt}\renewcommand{\arraystretch}{1}    \resizebox{\columnwidth}{!}{%
    \centering
    \setlength{\tabcolsep}{30pt}
    \begin{tabular}{|c|c|c|}
    \hline
    $\#$ResBlock  & mKID $\times$100$\downarrow$ &  mFID $\downarrow$  \cr
    \hline
    % \multirow{}{*}{}
       1   & 9.48	& 115.47	 \cr\cline{1-3}\hline
   2    & 9.39	& 110.31	 \cr\cline{1-3} \hline
        3    & 9.34 & 105.98 \cr\cline{1-3}\hline
    4    & 9.25 & 103.55	 \cr\cline{1-3}\hline
    \end{tabular}   
    }
\caption{\small Ablation on number of the shared layers between the generator $G$ and the \textit{auxiliary generator} $G'$. Note we we account for the shared ResBlock  layer from the bottom layer of the generator. \vspace{-7mm} }\label{tab:abalation_layer_aux}
\end{table}

\minisection{The baselines for multi-class I2I.}
We compare to StarGAN~\cite{StarGAN2018}, StarGANv2~\cite{choi2020stargan}, SDIT~\cite{wang2019sdit}, DRIT++~\cite{lee2020drit++}, DMIT~\cite{yu2019multi} and DeepI2I~\cite{wang2020deepi2i}, all of which perform image-to-image translation between multi-class domains. StarGANv2~\cite{choi2020stargan} obtains the scability by introducing a class-specific network.
\textit{SDIT~\cite{wang2019sdit}} leverages the class label and random noise to achieve  
 scalability and diversity in a single model. A similar idea is also explored in \textit{DMIT~\cite{yu2019multi}}.

\begin{figure}[t]
    \centering
    \includegraphics[width=\columnwidth]{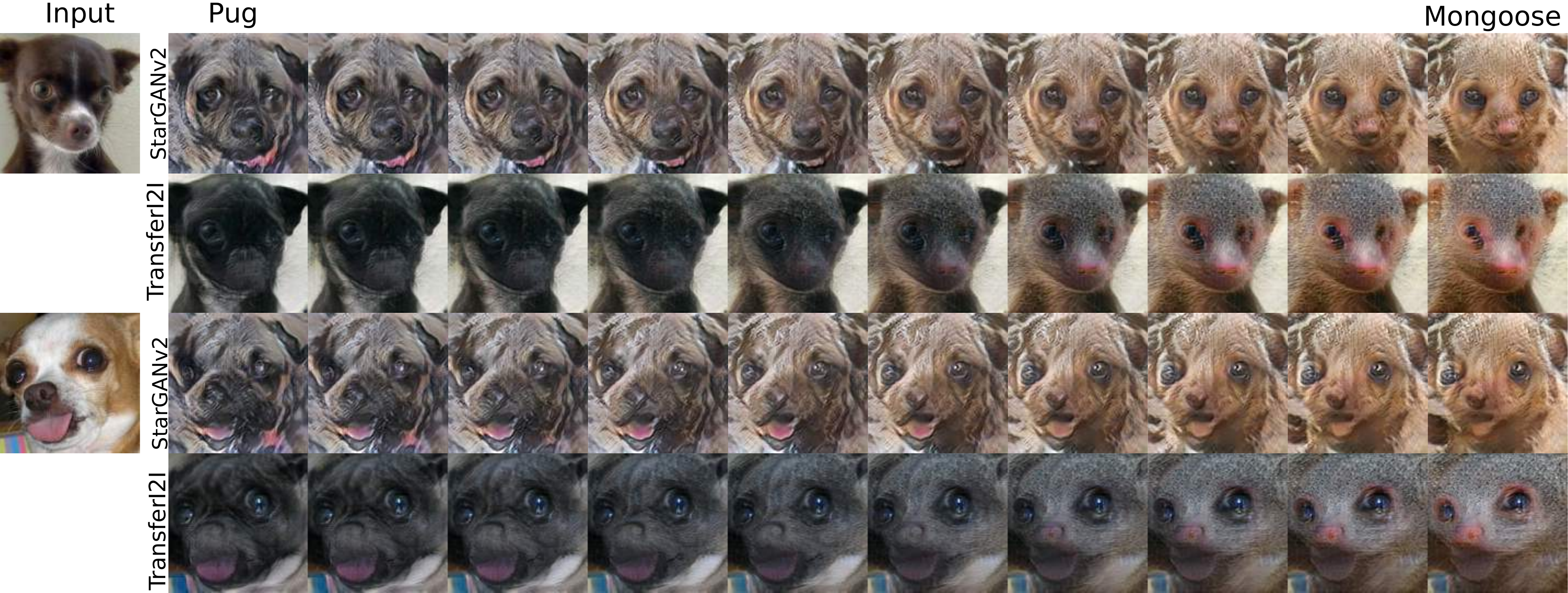}
    \caption{   \small Interpolation by keeping the input image  fixed while interpolating between two class embeddings. The first column is the input images, while the remaining columns are the interpolated results. The interpolation results from \textit{pug} to \textit{mongoose}.
    }\vspace{-4mm}
    \label{fig:animal_interpolation}
\end{figure}

\begin{figure*}[t]
    \centering
    \includegraphics[width=0.95\textwidth]{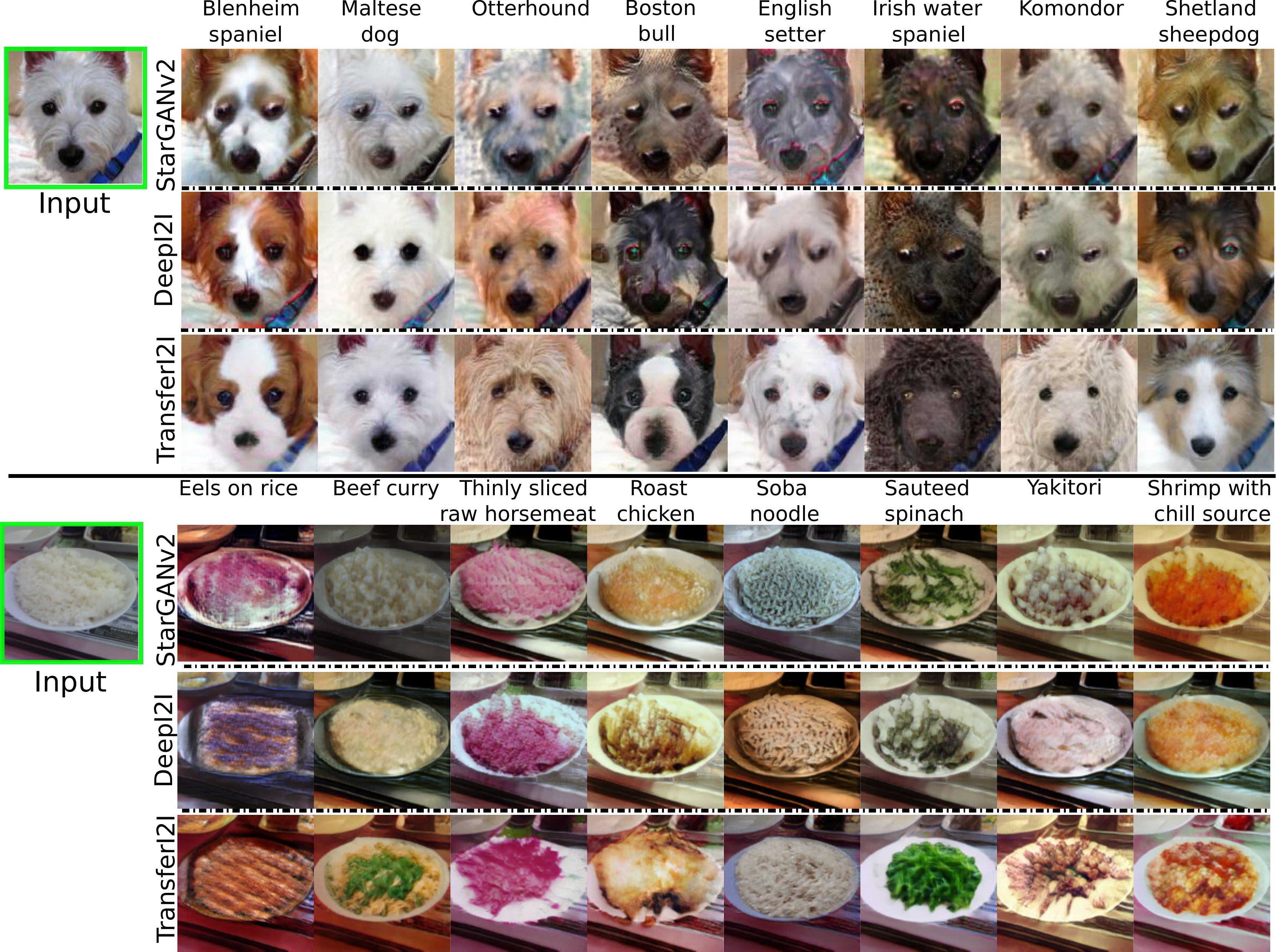} 
    \caption{\small Qualitative comparison on the \textit{Animal faces} and \textit{Foods}. The input images are in the first column and the remaining columns show the class-specific translated  images.  } 
    \label{fig:compare_baseline}\vspace{-3mm}
\end{figure*}

\begin{table*}[t]
    \setlength{\tabcolsep}{1mm}
    \resizebox{\textwidth}{!}{%
    \centering
    \footnotesize
    \setlength{\tabcolsep}{5pt}
    \begin{tabular}{|c|c|c|c|c|c|c|c|c|c|c|c|c|}
    \hline
    \multirow{2}{*}{\diagbox{Method}{Datasets}}& 
    \multicolumn{4}{c|}{\textit{Animal faces} (\textit{10/per class})}&\multicolumn{4}{c|}{\textit{Birds} (\textit{78/per class})}&\multicolumn{4}{c|}{\textit{Foods} (\textit{110/per class})}\cr\cline{2-13}
       & mKID$\times$100$\downarrow$   & mFID$\downarrow$ & RC$\uparrow$   & FC$\uparrow$   &  mKID$\times$100$\downarrow$   & mFID$\downarrow$& RC$\uparrow$   & FC$\uparrow$   &  mKID$\times$100 $\downarrow$  & mFID$\downarrow$& RC$\uparrow$   & FC$\uparrow$  \cr
    \hline 
      StarGAN   	&28.4	&276.5 &4.89	&5.12& 21.4&214.6&9.61	&10.2& 20.9&210.7&10.7	&12.1 \cr\cline{1-13}
      SDIT  	&31.4	&283.6&5.51	&4.64 & 22.7&223.5&8.90	&8.71	& 23.7&236.2 &11.9	&11.8\cr\cline{1-13}
       DMIT&29.6 &280.1 &5.98	&5.11 &23.5&230.4&12.9	&11.4& 19.5&201.4&8.30	&10.4\cr\cline{1-13}
       DRIT++  &26.6 &270.1&4.81	&6.15  & 24.1&246.2&11.8	&13.2& 19.1&198.5&10.7	&12.7\cr\cline{1-13}
       StarGANv2 & 11.38& 131.2&12.4	&14.8 & 10.7&152.9&25.7	&21.4&6.72&142.6&34.7	&22.8\cr\cline{1-13}
      TransferI2I (scratch)  &	41.37 & 356.1 &3.47	&1.54 & 30.5 & 301.7&3.24	&5.84&  26.5&278.2 &5.83	&4.67\cr\cline{1-13}
       DeepI2I &	11.48 &	137.1&10.3	&9.27 & 8.92 &146.3&20.8	&22.5& 6.38&130.8 &30.2	&19.3\cr\cline{1-13}
       TransferI2I &	\textbf{9.25} &	\textbf{103.5} &\textbf{22.3}	&\textbf{25.4}& \textbf{6.23} &\textbf{118.3}&\textbf{27.1}	&\textbf{28.4}& \textbf{3.62}&\textbf{107.8}&\textbf{43.2}	&\textbf{24.8} \cr\cline{1-13}
    \hline
    \end{tabular} 

    }
\caption{\small Comparison with baselines. TransferI2I obtains superior results on three datasets. We still obtain satisfactory advantage on both the bird and the food dataset, even they have more samples.}\vspace{-7mm}\label{tab:comparsion_baselines}
\end{table*}
%~

\begin{figure*}[t]
\centering
    \includegraphics[width=\textwidth]{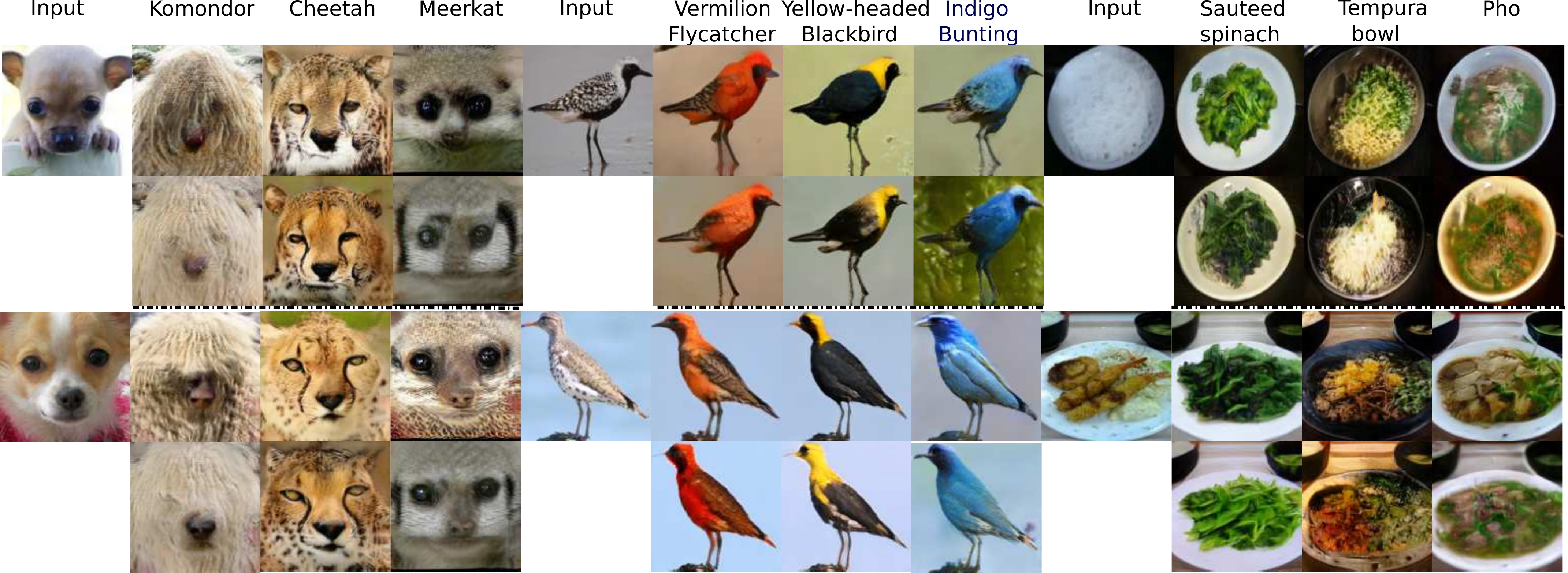}
\caption{\small Qualitative results of TransferI2I. The input image is in the first column and the class-specific outputs are in the remaining columns. For each specific target class, we show two images. }\vspace{-5mm}\label{fig:diversity}
\end{figure*}

\subsection{Multi-class I2I translation}\label{sec:many_class_case}

\minisection{Ablation study.} 
We now evaluate the effect of each independent contribution on the performance of TransferI2I. First we ablate the \textit{source-target initialization} and \textit{self-initialization} without the \textit{auxiliary generator}. Next, we evaluate the performance gain when adding the \textit{auxiliary generator}.

\minisection{\textit{Source-target initialization and \textit{self-initialization}.}} Table~\ref{tab:abalation_FT_AD_AU} reports the performance of both techniques in terms of both mFID and mKID on \emph{Animal faces}. 
Note that the second row of Table.~\ref{tab:abalation_FT_AD_AU} is equal to DeepI2I.
Adding one of the techniques (\emph{source-target initialization} and \textit{self-initialization} of the adaptor layer) improves performance of I2I translation compared to DeepI2I. % 
Furthermore, performing \textit{source-target initialization} achieves a larger advantage than \textit{self-initialization}, \textit{e.g.} for \textit{mFID}: 114.23 vs. 122.12. This seems to indicate the former is more important. 
Finally, using both techniques obtains the best mFID score, indicating that our method successfully performs I2I translation with few images.

\minisection{\textit{Auxiliary generator.}} In this paper, we propose to leverage the \textit{auxiliary generator} to encourage the usage of the deep layers of the generator. 
We conduct an experiment to evaluate the effect of the number of shared layers between the generator $G$ and the \textit{auxiliary generator} $G'$. As reported in Table.~\ref{tab:abalation_layer_aux}, we  found that more shared layers result in better performance (e.g., the mFID value reduces with an increasing number of shared layers). 

To measure the distance between two models we use weight fluctuation (WF), defined for two  models with parameters $\theta_1$ and $\theta_2$ as $\textrm{WF}=(\theta_1-\theta_2)^T \textrm{FM}_{\theta_1} (\theta_1-\theta_2)$ where $\textrm{FM}_{\theta_1}$ is the Fisher matrix~\cite{kirkpatrick2017overcoming}. This distance takes into account the importance of the weights to the loss computation. 
As shown in Figure~\ref{fig:ewc}, using the \textit{auxiliary generator} leads to larger weight changes in the deep layers than not using it, clearly demonstrating improved utilization and a beneficial effect in the overall system performance.   The drastic change (i.e., \textit{RB\_0 (w Aug.)} \textit{vs.} \textit{RB\_0 (w/o Aug.)}) appears in the first ResBlock of the generator, which means we are able to learn the semantic information. Moving towards the upper layers, the gap  of the corresponding layers of the two generators (w and w/o the \textit{auxiliary generator}) becomes smaller. The most likely reason is that the upper layers (influencing the structural information) use more information from the skip connections. 

\minisection{Interpolation.}
Figure~\ref{fig:animal_interpolation} reports interpolation by freezing the input images while interpolating the class embedding between two classes. Our model still manages to generate high quality images even for never seen class embeddings. On the contrary, StarGANv2 with limited data shows unsatisfactory  performance. %

\minisection{Quantitative results.}
As reported in Table~\ref{tab:comparsion_baselines},  we compare the proposed method with the baselines on the \textit{Animal faces}~\cite{liu2019few}, \emph{Birds}~\cite{van2015building} and \emph{Foods}~\cite{kawano2014automatic} datasets. Our approach outperforms all baselines in terms of mFID/mKID (joint quality and diversity) and RC/FC (the ability to generate class-specific images). We obtain a drop in mFID of around 30 points for both \emph{Animal faces} and \textit{Birds}, and 23 points on \textit{Foods}. This indicates an advantage of the proposed method on small datasets (e.g., the number of images per class is 10 on \emph{Animal faces}). The advantage is less on larger datasets (e.g. on the \textit{Food} dataset with 110 images per class). Training the same architecture from scratch (\textit{Transfer(scratch)}) obtains inferior results. %
Both StarGANv2 and DeepI2I exhibit similar performance, albeit inferior to TransferI2I on all metrics. Apart from the improvement on mFID, also the classification scores RC and FC of TransferI2I
show that both quality (RC/FC) and diversity (FC) are improved.

\begin{table}[t]

    \setlength{\tabcolsep}{1mm}
    \resizebox{\columnwidth}{!}{%
    \centering
    \footnotesize
    \setlength{\tabcolsep}{1pt}
    \begin{tabular}{|c|c|c|c|c|c|c|c|c|c|c|}
    \hline
    \multirow{3}{*}{\diagbox{Method}{Dataset}} &\multicolumn{4}{c|}{(cat,dog):(100,100) }&\multicolumn{4}{c|}{(cat,dog):(771,1264)}\cr\cline{2-9}
    &\multicolumn{2}{c|}{dog $\rightarrow$ cat}&\multicolumn{2}{c|}{cat $\rightarrow$ dog}& 
    \multicolumn{2}{c|}{dog $\rightarrow$ cat}&\multicolumn{2}{c|}{cat $\rightarrow$ dog}\cr\cline{2-9}
      & FID$\downarrow$ & KID $\downarrow$ & FID$\downarrow$ & KID $\downarrow$ & FID$\downarrow$ & KID $\downarrow$ & FID$\downarrow$ & KID $\downarrow$  \cr\cline{2-9}
 \hline
    % \multirow{}{*}{
      CycleGAN   &210.7	&14.33	&284.6	&28.14&119.32	&4.93&125.30	&6.93\cr\cline{1-9}
      UNIT   &189.4	&12.29	&266.3	&25.51&59.56	&1.94&63.78	&1.94\cr\cline{1-9}
      MUNIT   &203.4	&13.63	&270.6	&26.17&53.25	&1.26&60.84	&7.25\cr\cline{1-9}
      NICEGAN   & 104.4	&6.04	&156.2	&10.56&48.79	&1.58&44.67	&1.20	 \cr\cline{1-9}
       UGATIT-light &133.3 &6.51 &206.6 &15.04 &80.70 &3.22 &64.36 &2.49  \cr\cline{1-9}
        CUT   &197.1	&12.01	&265.1	&25.83&45.41	&1.19&48.37	&6.37\cr\cline{1-9}
        StarGANV2   &336.4	&40.21	&339.8		&41.32&\textbf{25.41}		&0.91&\textbf{30.1}		&\textbf{1.03}\cr\cline{1-9}
        DeepI2I   & 83.71	&4.26	&112.4	&5.67&43.23	&1.37&39.54	&1.04	 \cr\cline{1-9}
        TransferI2I   & \textbf{55.2}	&\textbf{3.97}	&\textbf{83.6}	&\textbf{4.59}&27.0	&\textbf{0.84}&37.13	&1.12	 \cr\cline{1-9}
    \hline
    \end{tabular}  

    }
\caption{\small   The metric results   on both \textit{cat2dog-200} and \textit{cat2dog-2035} datasets. Note we multiply 100 for \textit{KID}.}\vspace{-1mm}\label{tab:cat_dog_metric}
\end{table}

\minisection{Qualitative results.}  
Figure~\ref{fig:compare_baseline} shows the comparison to  baselines on \textit{Animal faces} and \textit{Foods} dataset.  Although both StarGANv2 and DeepI2I are able to perform multi-class I2I translation to each class, they fail to generate highly realistic images. Taking \emph{Animal faces} as an example, given the target class label our method is able to provide high visual quality images. The qualitative results of the \textit{Foods} dataset also confirm our conclusion: the images of TransferI2I are in general of higher quality than those of the baselines. 
We further validate whether our method has both scalability and diversity in a single model. As shown in Figure~\ref{fig:diversity}, given the target class label (e.g., \textit{Komondor}) our method successfully synthesizes diverse images by varying the noise $z$ (i.e., the second column of the figure). The results show that by changing the target class label (i.e., scalability) the generator produces the corresponding target-specific output.

\begin{figure}[t]
\centering
    \includegraphics[width=\columnwidth]{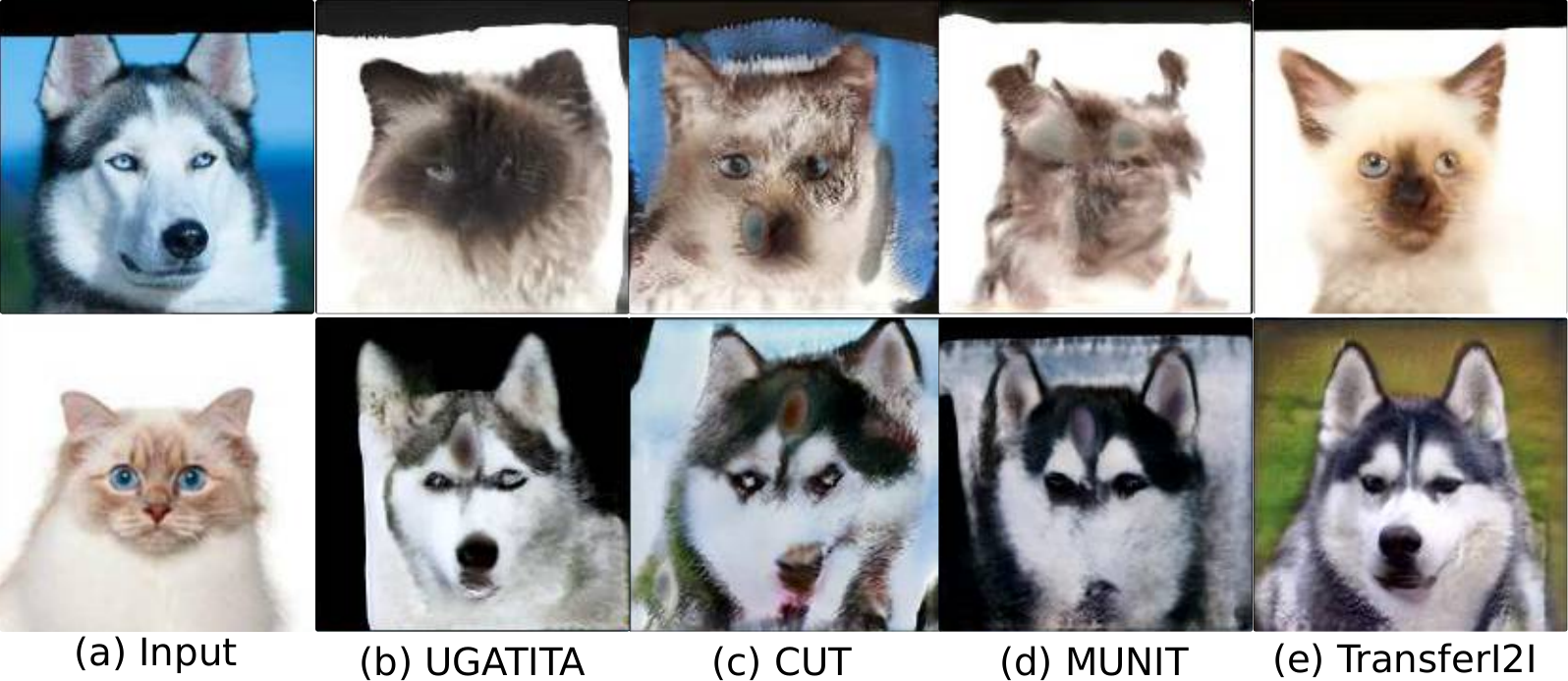}
\caption{\small Examples  of  generated  outputs   on \textit{cat2dog-200} dataset. }\vspace{-5mm}\label{fig:cat_dog_visualization}
\end{figure}

\subsection{Two-class I2I translation}\label{sec:two_class_case}
To evaluate the generality of our method, here we validate the proposed algorithm for two-class I2I translation on a two-category dataset: cats and dogs. We use the pretrained StyleGAN to initialize our model (see Figure~\ref{fig:models} (a)).   Figure~\ref{fig:cat_dog_visualization}  shows the generated image of both the baselines and the proposed method on \textit{cat2dog-200} dataset. We can easily observe that the baselines fail to synthesize realistic image, although they learn the style information of the target domain.  We can see that TransferI2I generates more realistic target images. For quantitative evaluation, we report the results in terms of FID and KID. As shown in  Table~\ref{tab:cat_dog_metric},  TransferI2I obtains the best score on the small \textit{cat2dog-200} dataset, improving FID by around 30 points with respect to DeepI2I. This clearly demonstrates that our method successfully conducts I2I translation when given limited data. On the much larger \textit{cat2dog-2035} dataset, where transfer learning is less crucial, we obtain comparable performance to StarGANv2 but significantly outperform DeepI2I (which uses a similar architecture).

\section{Conclusions}
We have proposed an approach to benefit from transfer learning for image-to-image methods.  We decoupled our learning process into an image generation and I2I translation step. The first step, including the \textit{source-target initialization} and \textit{self-initialization} of the adaptor, aims to learn a better initialization for the I2I translation (the second step). Furthermore, we introduce an \textit{auxiliary generator} to overcome the inefficient usage of the deep layers of the generator. Our experiments confirmed that the proposed transfer learning method can lead to state-of-the-art results even when training with very few labelled samples, outperforming recent methods like deepI2I and starGANv2.

\section*{Acknowledgements}
We acknowledge the support from Huawei Kirin Solution, the Spanish Government funding for projects PID2019-104174GB-I00 (AEI / 10.13039/501100011033), 
% yaxing
RTI2018-102285-A-I0, and the CERCA Program of the Generalitat de Catalunya.

\appendix

\section{Multi-class I2I translation }
Here we introduce how to perform unpaired multi-class I2I translation.   We consider two domains: source domain $\mathcal{X}_1 \subset \mathbb{R}^{H\times W\times 3} $ and target domain $\mathcal{X}_2 \subset \mathbb{R}^{H\times W\times 3} $ (it can trivially be extended to multiple classes). In this work, given limited training samples from both source and target domains, we aim to map a source image  $\mathbf{x}_1  \in \mathcal{X}_1 $ into a target sample $\mathbf{x}_{1 \rightarrow 2} \in \mathcal{X}_2$ conditioned on the target domain label $\mathbf{c}\in\left\{1,\ldots,C\right\}$ and a random noise vector $\mathbf{z \in \mathbb{R}^{Z}}$.   Let  image $ \mathbf{x} \in  \mathcal{X}_1 \bigcup \mathcal{X}_2 $ is sampled from dataset. 

\begin{figure*}[t]
    \centering
    \includegraphics[width=\textwidth]{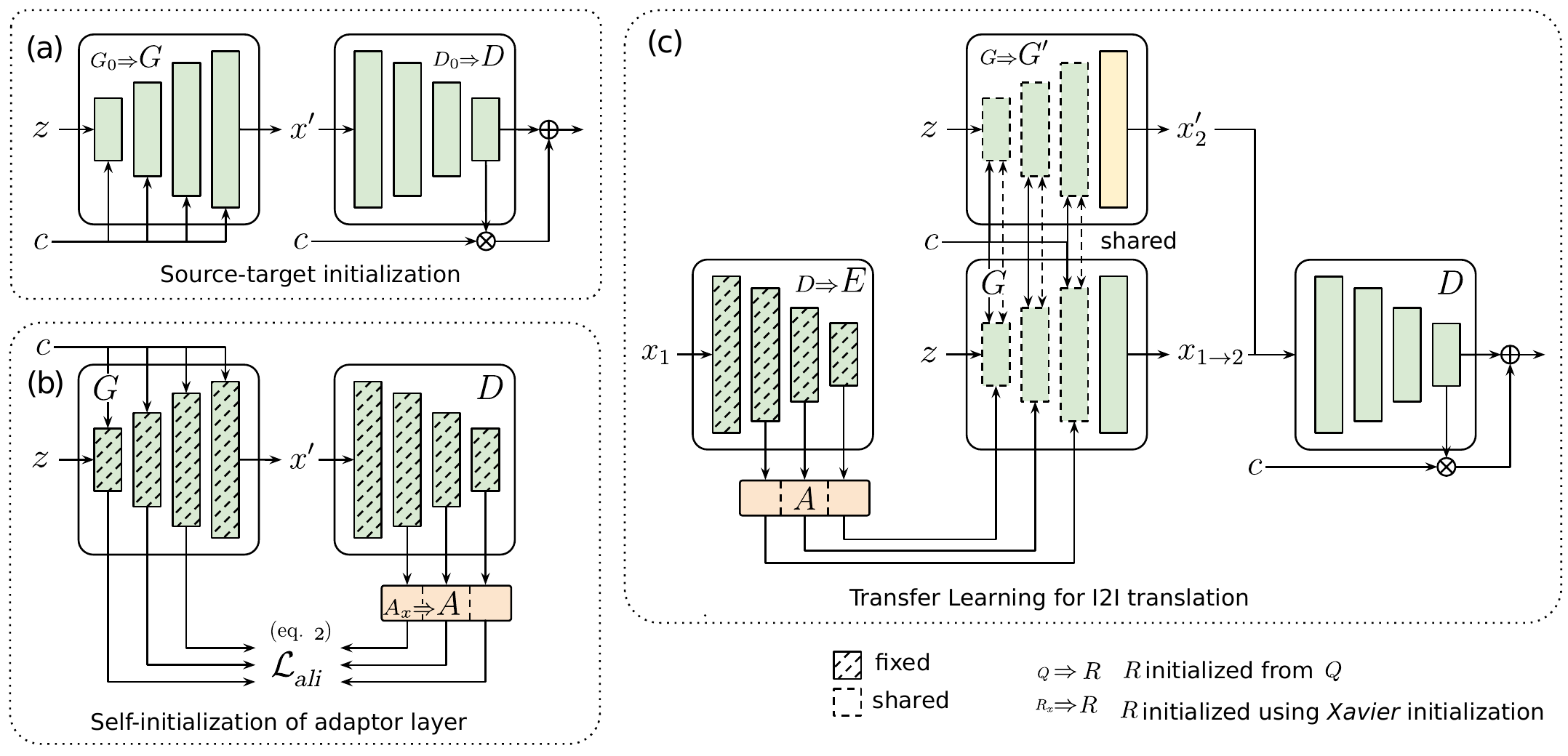}
    \caption{\small    \small Conditional model architecture and training stages. Here modules come from the immediate previous stage unless otherwise indicated. A pretrained GAN (e.g., BigGAN~\cite{brock2018large}) is used as $G_0$ and $D_0$ to initialize the  GAN. (a) \textit{Source-target initialization} performs finetuning on all data to form  a trained GAN model (i.e., the generator $G$ and the discriminator $D$). (b) \textit{Self-initialization} of adaptor layer to pretrain the adaptor $A$ and align both the generator $G$ and the discriminator $D$.   We only update the adaptor layers $A$. (c) The I2I translation model is composed of five main parts: the encoder $E$, the adaptor layer $A$, the generator $G$, the \textit{auxiliary generator} $G'$ and the  discriminator $D$. Note the encoder $E$ is initialized by the discriminator $D$. The portion of weights from $G'$ that is not shared (in yellow), is initialized with $G$ weights.
    }\vspace{-3mm}
    \label{supp:models}
\end{figure*}

As illustrated Figure~\ref{supp:models}, our framework is composed of three stages: \textit{source-target initialization} (Figure~\ref{supp:models}(a)) aiming to obtain a satisfactory domain-specific GAN, which can then be used for I2I translation;  \textit{self-initialization of adaptor layer} (Figure~\ref{supp:models}(b)) which reduces the risk of overfitting of the adaptor layers when trained on limited data; and \textit{transfer learning for I2I translation} (Figure~\ref{supp:models}(c)) which finetunes all networks, each of which is initialized according to the previous steps, on the few available source and target images.

\minisection{Source-target initialization.}  we expect to study a excellent generative model utilizing the limited training data. Different to the model for two-class I2I translation, in this stage we  train one generator and one discriminator on all images instead of class-specific generator and class-specific discriminator. The training objective is as following:

\begin{equation}
\begin{aligned}
\mathcal{L}_{GAN} &= \mathbb{E}_{\mathbf{x}\sim \mathcal{X}_1 \bigcup \mathcal{X}_2 }\left[ \log D \left ( \mathbf{x}, c \right ) \right]  \\
&+  \mathbb{E}_{\mathbf{z} \sim p(\mathbf{z}), \mathbf{c} \sim p(\mathbf{c})}\left[ \log \left ( 1 - D \left ( G \left ( \mathbf{z}, c \right ), c \right ) \right )   \right],
\end{aligned}
\end{equation} 
where $\mathbf{p} \left ( \mathbf{z} \right )$ follows the normal distribution, and $\mathbf{p} \left ( \mathbf{c} \right )$ is the domain label distribution. Here the generative model is used to provide a better initialization for the I2I translation.

\minisection{Self-initialization of adaptor layer.} We expect to overcome the overfitting of the adaptor layers, as well as  aligning the distribution of both the pretrained generator and the pretrained discriminator. As introduced in Section 3.1,  we propose the \textit{self-initialization} procedure, which leverages the previous pretrained model (Figure~\ref{supp:models} (a)) to achieve this goal. Especially,  both the noise $\mathbf{z}$ and the class embedding $\mathbf{c}$ are taken as input for the generator $G$, from which we extract the hierarchical representation $F_{g}(\mathbf{z}, \mathbf{c})  = \left \{ G(\mathbf{z}, \mathbf{c})_{l} \right \}$ as well as  the synthesized image $G(\mathbf{z},\mathbf{c})$. Here $G(\mathbf{z}, \mathbf{c})_{l}$ is the $l_{th} (l = m,\ldots,n , (n>m))$ ResBlock~\footnote{After each ResBlock the feature resolution is half of the previous one in both encoder and discriminator,  and  two times in generator} output of the generator $G$. We then take the generated image $G(\mathbf{z}, \mathbf{c})$ as input for the discriminator $D$, and similarly  collect the hierarchical feature  $F_{d}(\mathbf{z}) = \left \{ D(G(\mathbf{z}, \mathbf{c}))_{l} \right \}$. The adaptor network $A$ finally takes the output representation  $\left \{ D(G(\mathbf{z},\mathbf{c}))_{l} \right \}$ as input, that is $A(F_{d}(\mathbf{z})) = \left \{ A \right \} $. In this step, our loss is:
\vspace{-3pt}
\begin{equation}
\begin{aligned}
 \mathcal{L}_{ali}= \sum _{l}  \left \| F_g \left ( \mathbf{z} \right ) - A \left ( D(G(\mathbf{z}, \mathbf{c})) \right )  \right \|_{1}.
\end{aligned}
\end{equation}

\minisection{Transfer Learning for I2I translation.} Figure~\ref{supp:models}(c) shows how to map the image from the source domain to target domain. In this stage, we propose an \emph{auxiliary generator} $G'$ which aims to improve the usage of the deep layers of the generator, largely due to the skip connections.  It is relatively easy for the generator to use the information from the high-resolution skip connections (connecting to the upper layers of the generator), and ignore the deep layers of the generator, which require a more semantic understanding of the data, thus more difficult to train. 

Our loss function for I2I translation is a multi-task objective comprising: (a) \textit{conditional adversarial loss} which not only classifies the real image and the generated image, but encourages the networks $\{E, A, G\}$ to generate class-specific images which correspondent to label $\mathbf{c}$. (b) \textit{reconstruction loss}  guarantees that both the input image $\mathbf{x}_1$ and the synthesized image $\mathbf{x}_{1 \rightarrow 2} = G(\mathbf{z}, \mathbf{c}, A(E(\mathbf{x}_1)))$ keep the similar structural information.

\minisection{\textit{Conditional adversarial loss.}} We employ GAN~\cite{goodfellow2014generative}  to optimize this problem as follows:
\begin{equation}
\begin{aligned}
&\mathcal{L}_{GAN} = \mathbb{E}_{\mathbf{x}_2\sim \mathcal{X}_2,\mathbf{c} \sim p(\mathbf{c})}\left[ \log D \left ( \mathbf{x_2}, \mathbf{c} \right ) \right]  \\
&+   \mathbb{E}_{\mathbf{x}_1 \sim \mathcal{X}_1, \mathbf{z} \sim p(\mathbf{z}), \mathbf{c} \sim p(\mathbf{c})}\left[ \log (1 - D \left ( G \left ( A \left ( E \left ( \mathbf{\mathbf{x}_1} \right ) \right ), \mathbf{z}, \mathbf{c} \right )  \right ) \right]  \\
&+ \lambda_{aux} \mathbb{E}_{\mathbf{z} \sim p(\mathbf{z}), \mathbf{c} \sim p(\mathbf{c})}\left[ \log \left ( 1 - D \left ( G' \left ( \mathbf{z}, \mathbf{c}  \right ) \right ) \right )   \right],
\end{aligned}
\end{equation} 
The hyper-parameter $\lambda_{aux}$ balances the importance of each terms. We set $\lambda_{aux} = 0.01$.

\minisection{\textit{Reconstruction loss.} } We use reconstruction  to preserve the structure of both the input image $x_1$  and the output image $x_{1 \rightarrow 2}$. In the same fashion  as results for photo-realistic image generation~\cite{jakab2018unsupervised,johnson2018image,shrivastava2017learning}, we  use the discriminator output to achieve this goal through the following loss:

\begin{equation}
\begin{aligned}
 \mathcal{L}_{rec} &= \sum _{l} \alpha_{l} \left \lVert D \left ( \mathbf{x_1} \right ) - D \left ( \mathbf{x_{1 \rightarrow 2}} \right )  \right \rVert_{1} ,
\end{aligned}
\end{equation}
where parameters $\alpha_{l}$ are scalars which balance the terms. Note we set $\alpha_{l} = 1$.

\minisection{\textit{Full Objective.}}
The full objective function of our model is:
 \vspace{-2pt}
\begin{equation}
%\label{eq:loss}
\begin{aligned} 
&\min_{\substack{E, A, G, G^{'}}} \underset{D}{\max}   \mathcal{L}_{GAN} +  \lambda_{rec}  \mathcal{L}_{rec}
\end{aligned}
\end{equation}
where $\lambda_{rec}$ is a hyper-parameter that balances the importance of each terms.  We set $\lambda_{rec} = 1$.

\begin{figure*}[t]
    \includegraphics[width=\textwidth]{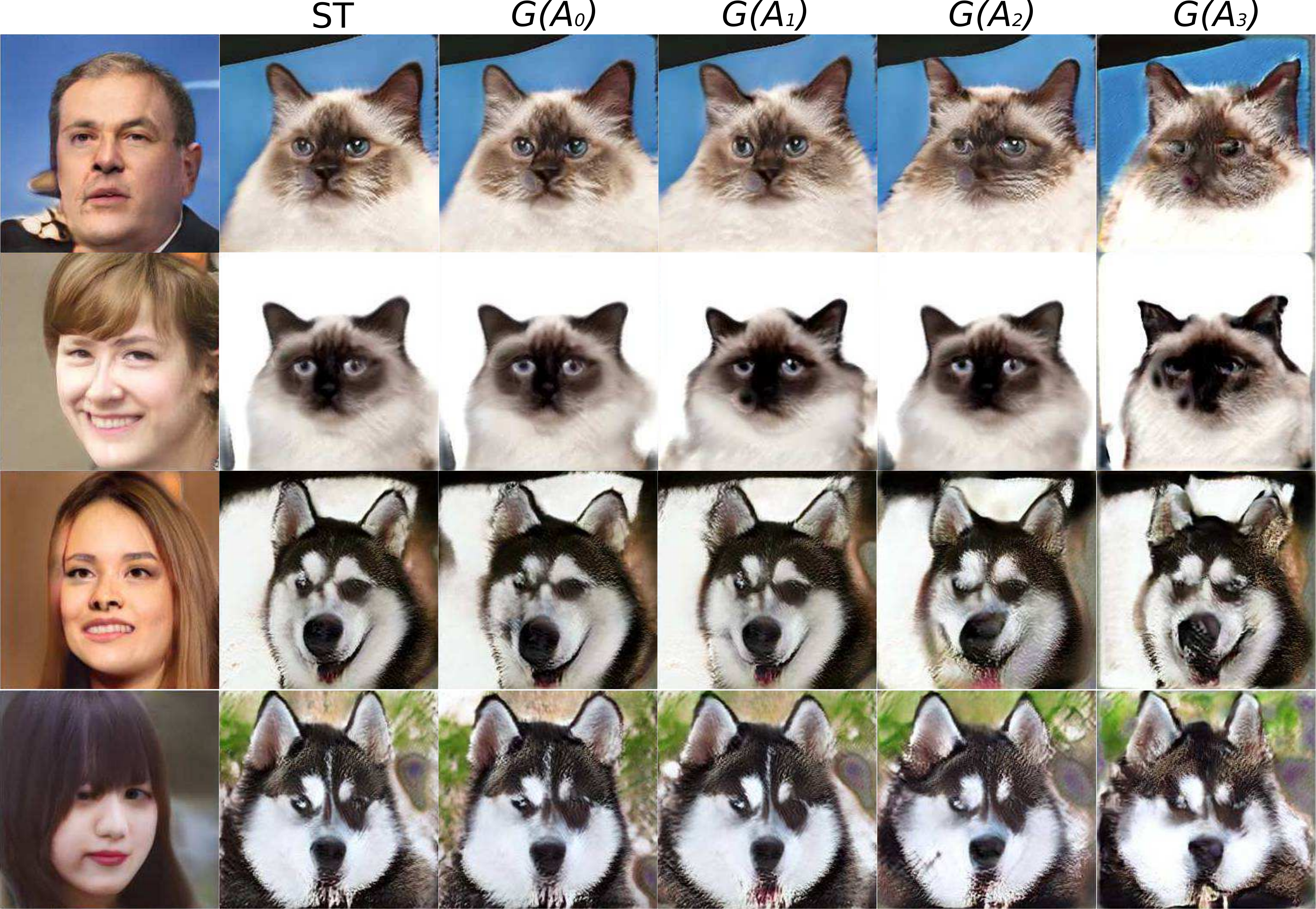}
\caption{\small Examples generated by both \textit{source and target initialization} and \textit{self-initialization} of the adaptor on \textit{cat2dog} dataset.  The first two columns are the output of the StyleGAN and the generator after the \textit{source and target initialization} respectively. The remaining columns ($G(A_i) (i = 0,1,2,3)$)  are the corresponding output of the generator $G$ which only takes the corresponding output of the adaptor $A_i (i = 0,1,2,3)$.}\label{fig:self_supervised}
\end{figure*}

\begin{figure*}[t]
    \includegraphics[width=\textwidth]{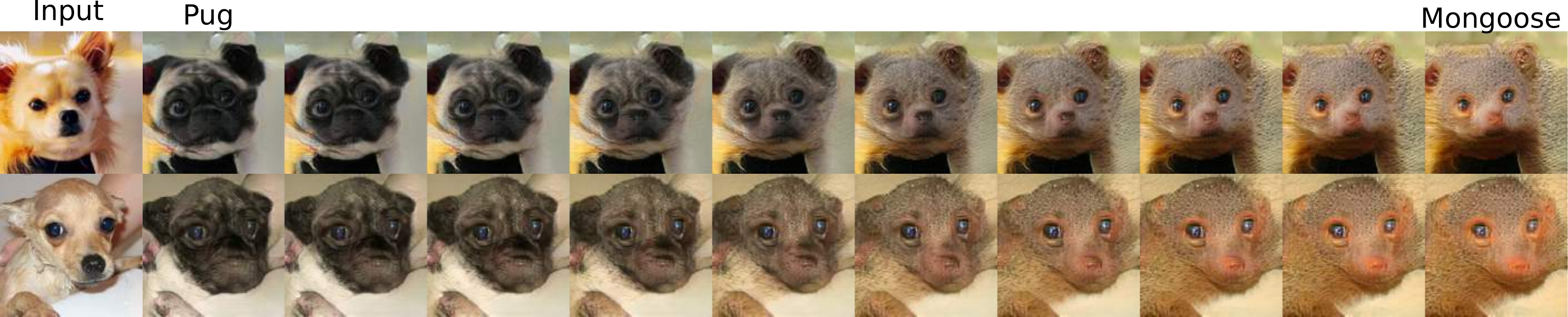}
\caption{\small Interpolation by keeping the input image  fixed while interpolating between two class embeddings. The first column shows the input images, while the remaining columns are the interpolated results. The interpolation results from \textit{pug} to \textit{mongoose}. }\label{fig:interpolation}
\end{figure*}

\section{Adaptor} \label{supp:adaptor}
We use the adaptor $A$ to connect the encoder $E$ and the generator $G$, aiming to leverage both the structure and semantic information. We sum the output of the adaptor with the corresponding one of the generator, which  is as following:

\begin{equation}
\begin{aligned}
 \hat{G}_{l}=  G_{l}\left(\mathbf{x}_1, \mathbf{z}, \mathbf{c}\right) + w_{l} A_{l}\left ( E_{l} \left ( \mathbf{x}_1 \right ) \right )
\end{aligned}
\end{equation}
where $G_{l}$ is the output of the corresponding layer which has same resolution to $A_{l}$. The hyper-parameters $w_{l}$ are used to balance the two terms (in this work we set $w_{l}$ is 1 except for the feature (32*32 size) which is 0.1 ).  Note for two-class I2I translation, we perform similar procedure.  
\section{Ablation study}
We further qualitatively compare the generated images after \textit{source and target initialization} on two-class I2I translation. The second column of Figure~\ref{fig:self_supervised} shows the synthesized images after \textit{source and target initialization}. We can see that the produced images are highly realistic and category-specific, indicating the effectiveness of this step. Next, we want to verify whether the \textit{self-initialization} of the adaptor successfully aligns encoder and generator. Therefore,  we take the noise as input for the generator, and  obtain an image, which is further fed into the discriminator and then through the adaptor layer. The adaptor layer output is then used as the only input of the generator (now no noise input $z$ is given). The results are provided in the third to last columns of Figure~\ref{fig:self_supervised}. The generator still produces high fidelity images when only inputting the output features from the adaptor. These results demonstrate that the distribution of the adaptor is aligned to the generator before performing the transfer learning for I2I translation (Figure~\ref{supp:models}(c)).

\section{Results}

\begin{figure*}[t]
    \centering
    \includegraphics[width=\textwidth]{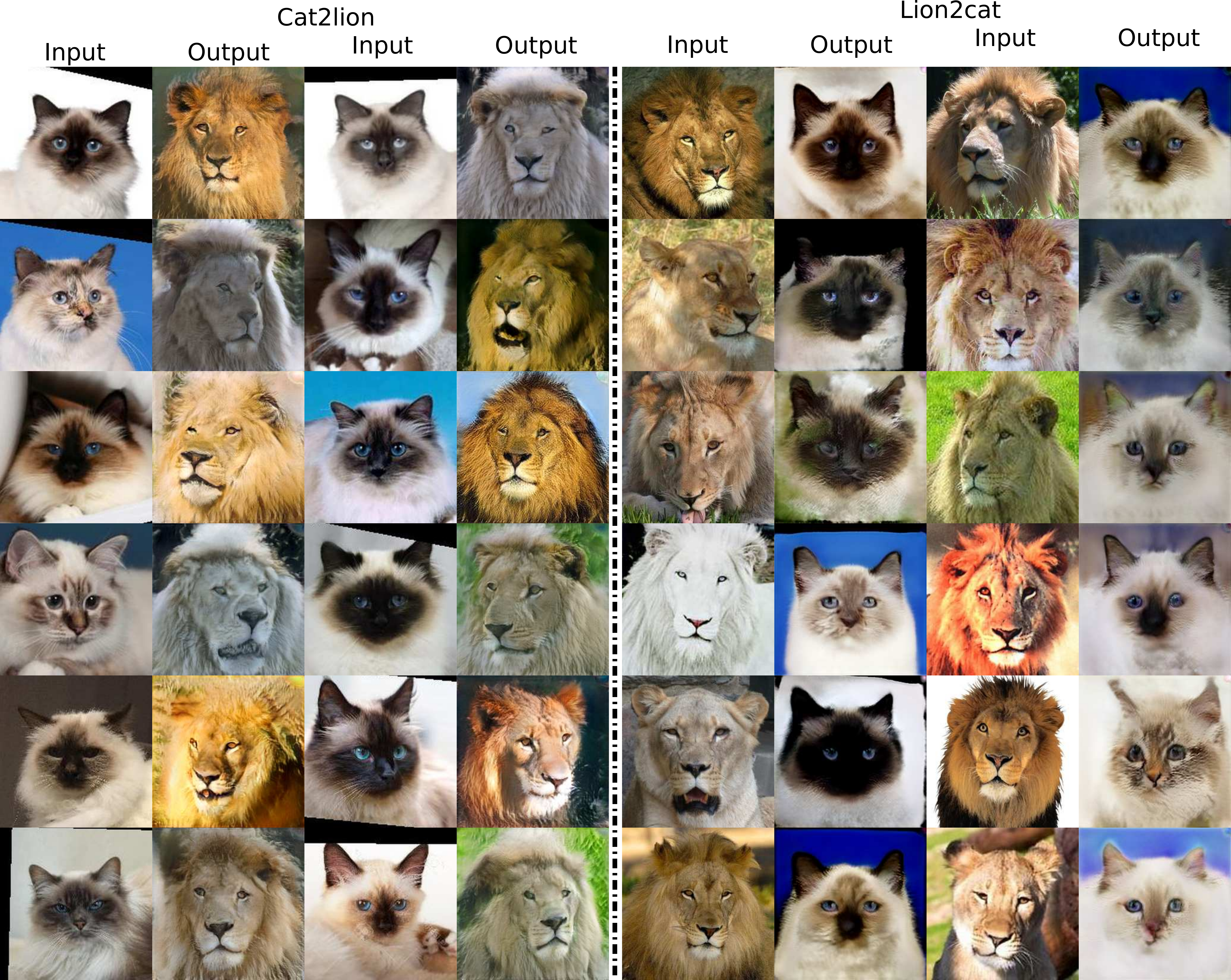}
    \caption{\small Qualitative results on both \textit{cat2lion} and \textit{lion2cat} dataset.  \vspace{-4mm}
    }

    \label{supp:cat2lion}
\end{figure*}

We provide additional results for the interpolation in Figure~\ref{fig:interpolation}.
We  evaluate the proposed method  on both \textit{cat2lion}  and \textit{lion2cat} datasets, which has 100 images for each category. The qualitative results are shown in Figure~\ref{supp:cat2lion}.

\begin{figure*}[t]
    \centering
    \includegraphics[width=\textwidth]{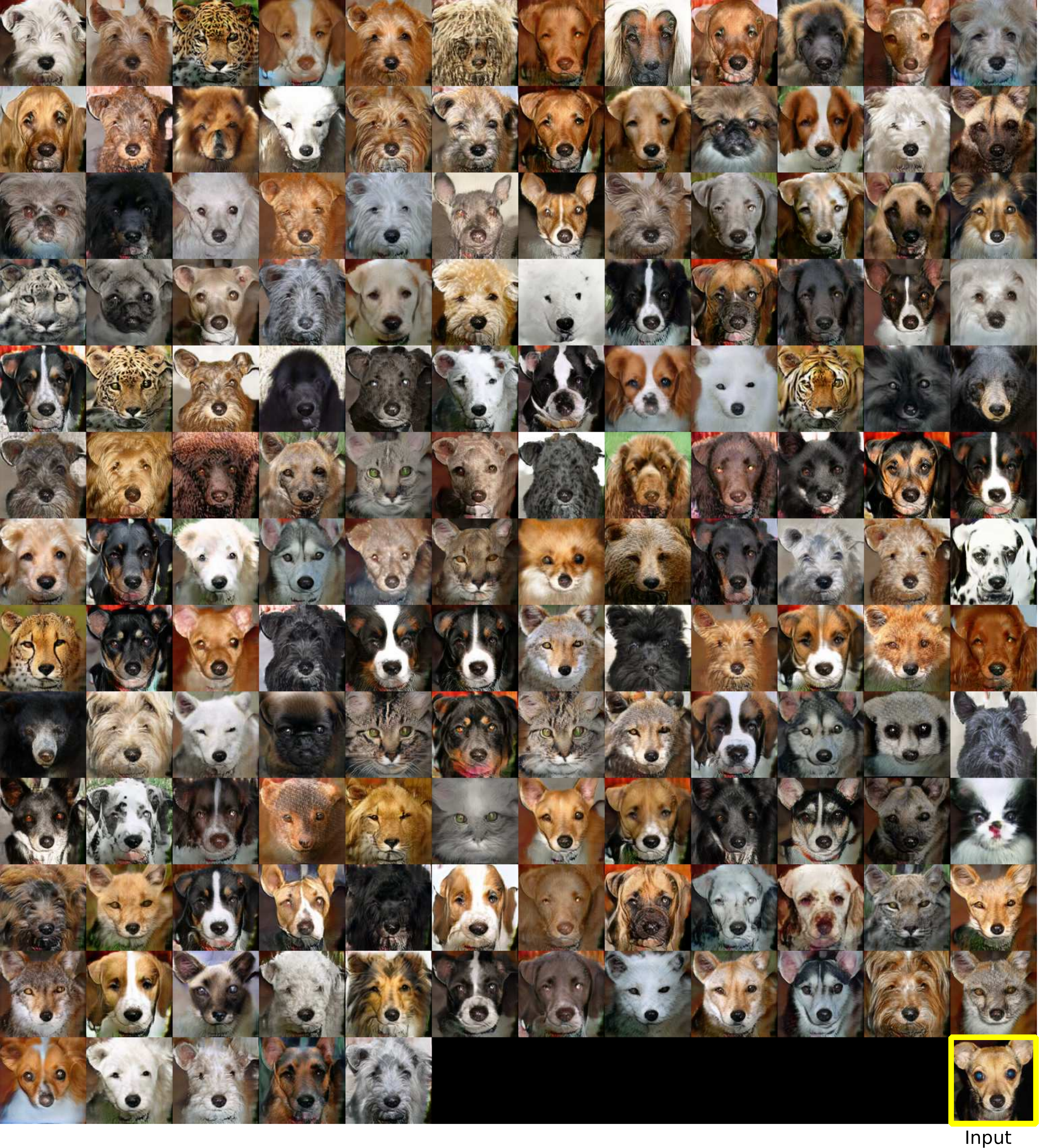}
    \caption{\small Qualitative results on the \textit{Animal faces} dataset. We translate the input image (bottom right) into all 149 categories. Please zoom-in for details. \vspace{-4mm}
    }

    \label{supp:animals_all_classes}
\end{figure*}

\begin{figure*}[t]
    \centering
    \includegraphics[width=\textwidth]{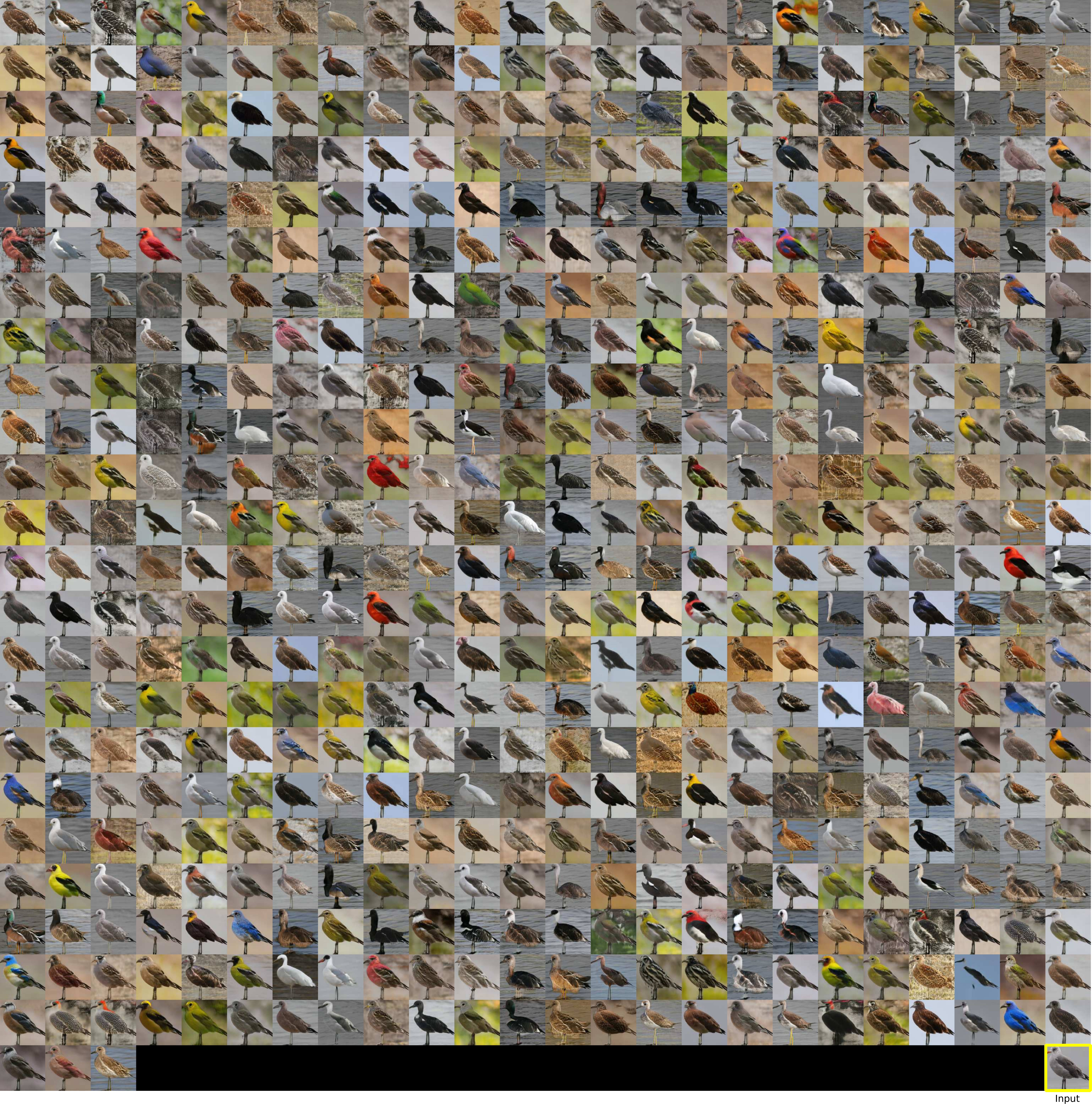}
    \caption{\small Qualitative results on the \textit{Birds} dataset. We translate the input image (bottom right) into all 555 categories. Please zoom-in for details. \vspace{-4mm}
    }

    \label{supp:birds_all_classes}
\end{figure*}

We  also show results translating an input image into all category on the \textit{Animal faces}, \textit{Foods}, and \textit{Birds} in Figure~\ref{supp:animals_all_classes}, and \ref{supp:birds_all_classes}.

\section{T-SNE}
\label{t-nse}
\begin{figure*}[t]
    \centering
    \includegraphics[width=\textwidth]{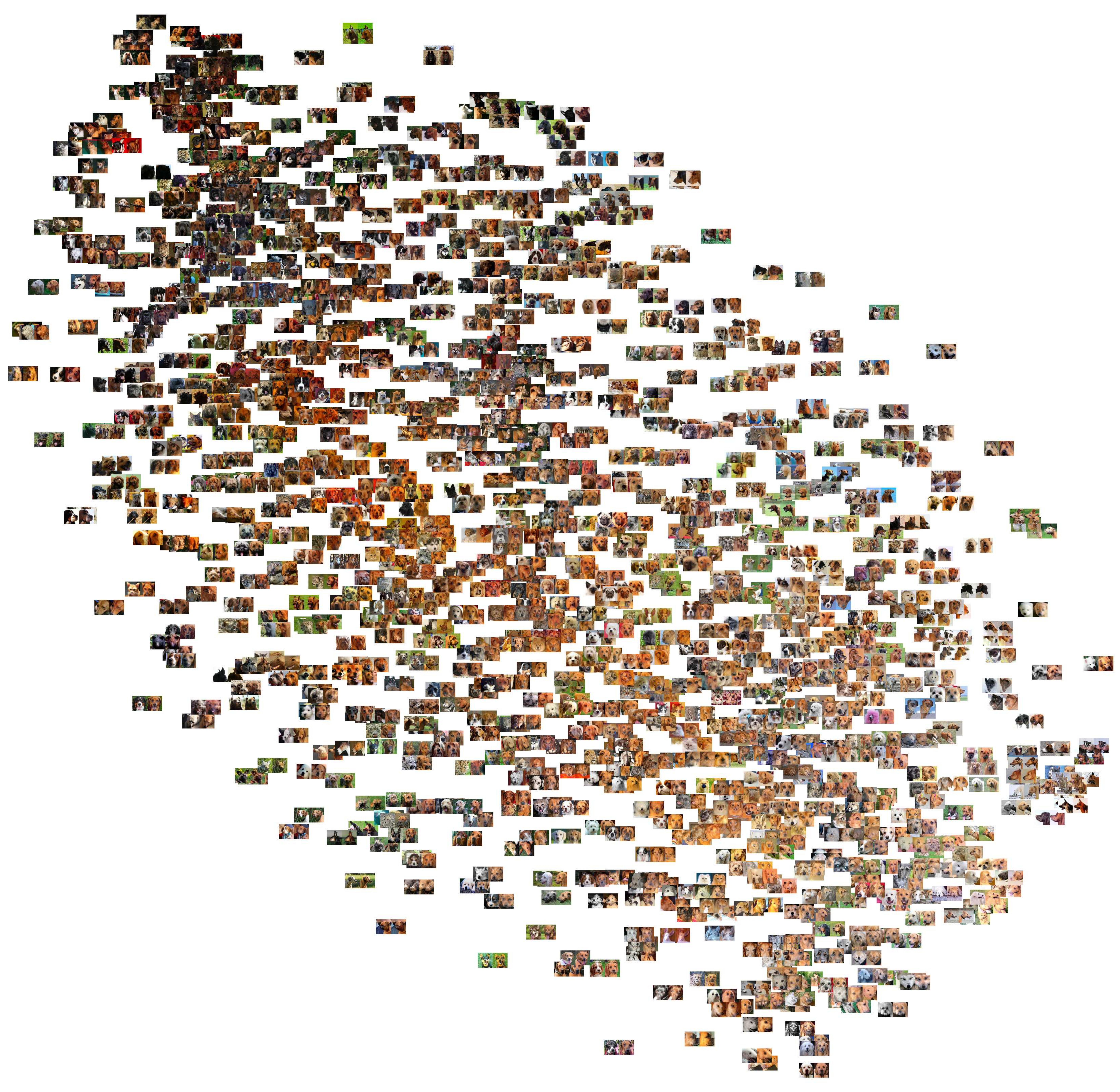}
    \caption{\small 2-D representation of the  T-SNE for 1280 generated images, the target class is \textit{Rhodesian ridgeback}. Note that for each pair image, the left is the input and the right is the output image. Please zoom-in for details. \vspace{-4mm}
    }

    \label{supp:tsne_single_class}
\end{figure*}

\begin{figure*}[t]
    \centering
    \includegraphics[width=\textwidth]{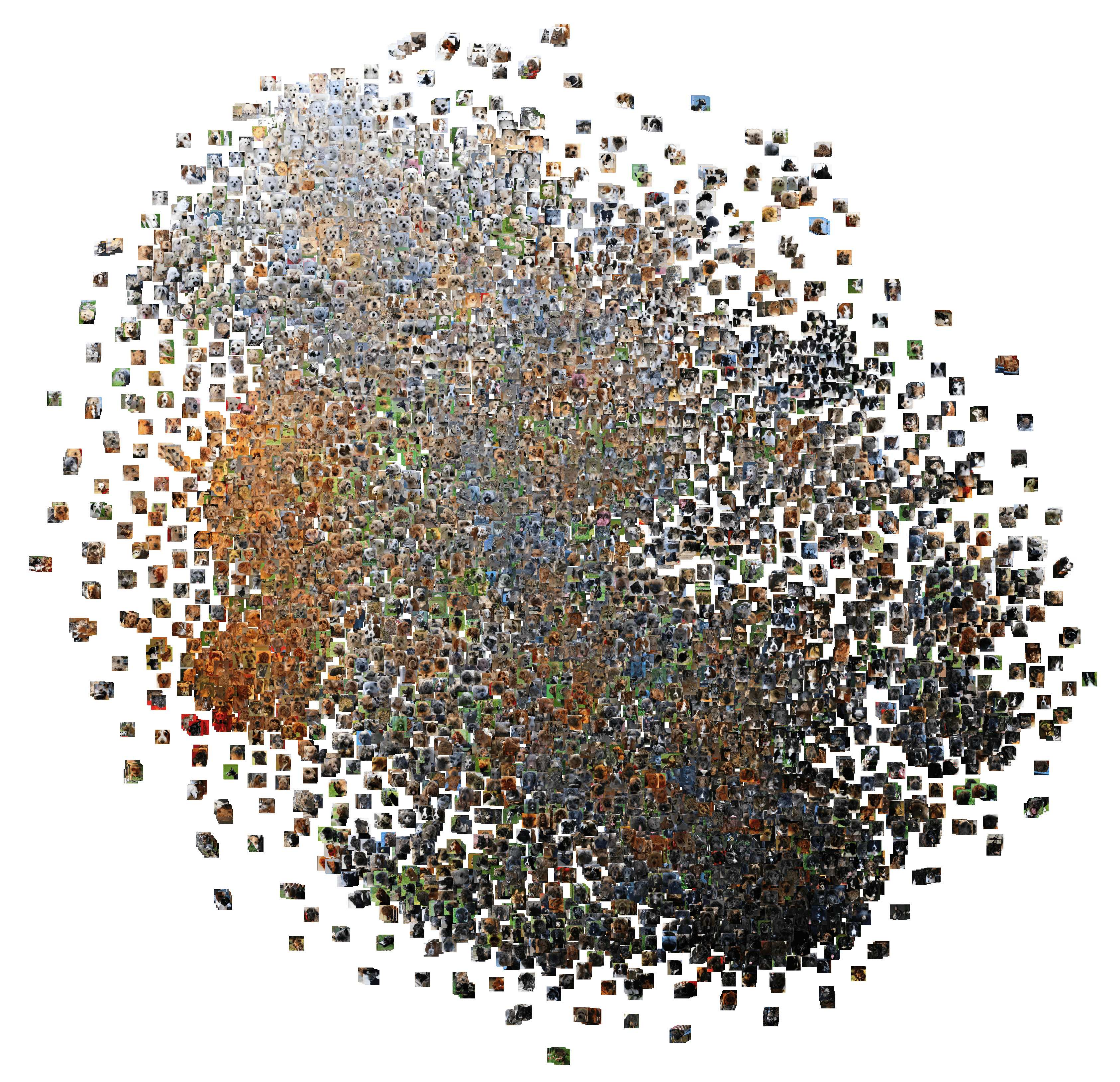}
    \caption{\small 2-D representation of the T-SNE for 14900 generated images across 149 classes. Please zoom-in for details. \vspace{-4mm}
    }

    \label{supp:tsne_many_class}
\end{figure*}

We explore  the latent space of the generated images. Given the target class $\mathbf{c}$ (e.g., \textit{Rhodesian ridgeback})), we take  different noises $\mathbf{z}$ and  the constant $\mathbf{c}$ as input for the networks $\{E, A, G \}$, and generate 1280 images.  Thus we use Principle Component Analysis (PCA)~\cite{bro2014principal} to extracted feature, following  the T-SNE~\cite{maaten2008visualizing} to visualize the generated images in a two dimensional space.
 As shown in Figure~\ref{supp:tsne_single_class}, given the target class (Rhodesian ridgeback), TransferI2I correctly disentangles the pose information of the input classes. The T-SNE plot shows that input animals having similar pose are localized close to each other in the T-SNE plot. Furthermore, it shows TransferI2I has the ability of diversity.
We also conduct T-SNE for 14,900 generated images across 149 categories (Figure~\ref{supp:tsne_many_class}).

\begin{figure*}[t]
    \centering
    \includegraphics[width=\textwidth]{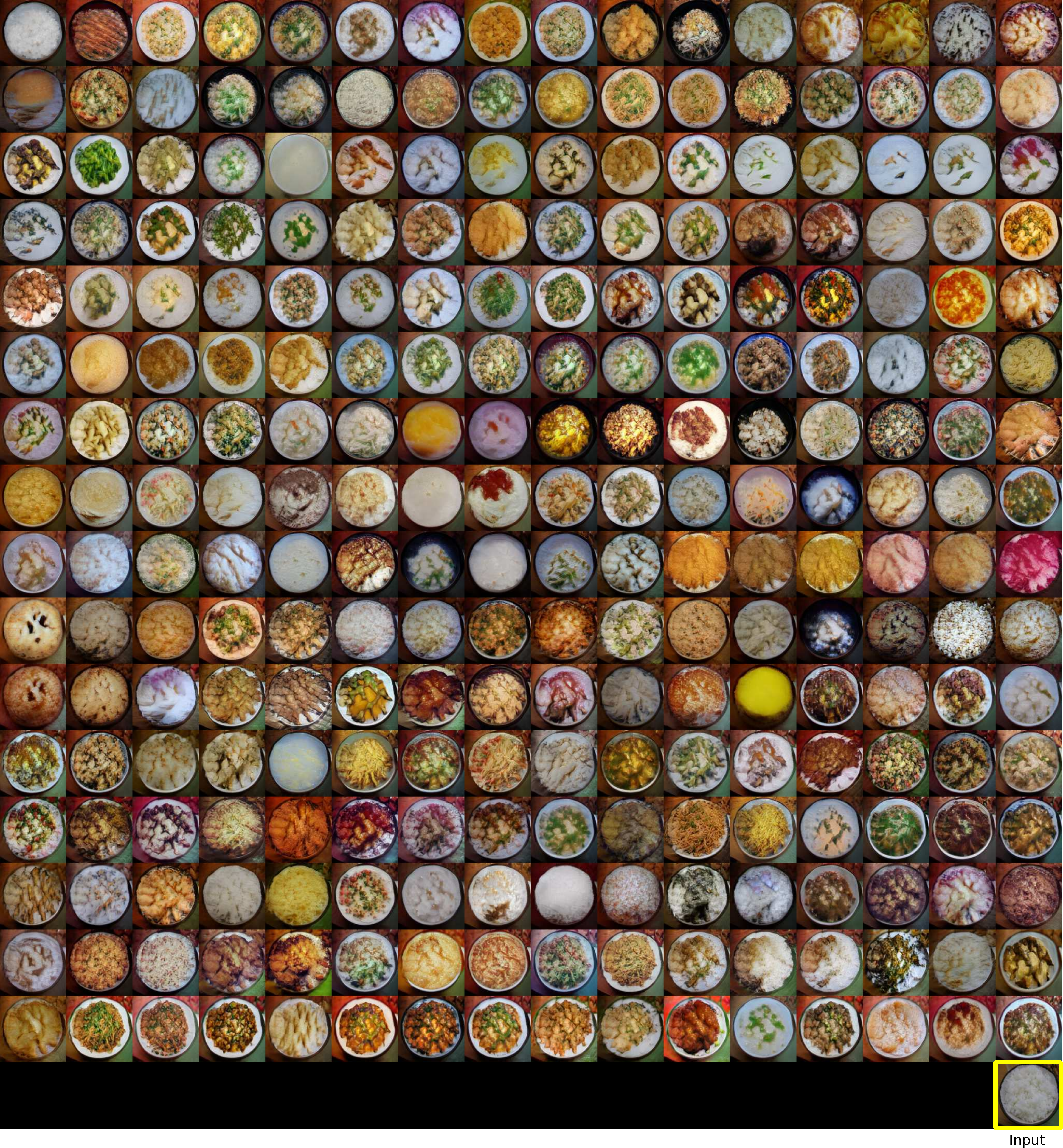}
    \caption{\small Qualitative results on the \textit{Foods} dataset. We translate the input image (bottom right) into all 256 categories. Please zoom-in for details. \vspace{-4mm}
    }

    \label{supp:foods_all_classes}
\end{figure*}

{\small
\bibliographystyle{ieee_fullname}
\bibliography{shortstrings,refs}

\begin{thebibliography}{10}\itemsep=-1pt

\bibitem{abdal2019image2stylegan}
Rameen Abdal, Yipeng Qin, and Peter Wonka.
\newblock Image2stylegan: How to embed images into the stylegan latent space?
\newblock In {\em ICCV}, pages 4432--4441, 2019.

\bibitem{almahairi2018augmented}
Amjad Almahairi, Sai Rajeswar, Alessandro Sordoni, Philip Bachman, and Aaron
  Courville.
\newblock Augmented cyclegan: Learning many-to-many mappings from unpaired
  data.
\newblock In {\em ICML}, 2018.

\bibitem{arjovsky2017wasserstein}
Martin Arjovsky, Soumith Chintala, and L{\'e}on Bottou.
\newblock Wasserstein gan.
\newblock In {\em ICLR}, 2017.

\bibitem{bau2018gan}
David Bau, Jun-Yan Zhu, Hendrik Strobelt, Bolei Zhou, Joshua~B Tenenbaum,
  William~T Freeman, and Antonio Torralba.
\newblock Gan dissection: Visualizing and understanding generative adversarial
  networks.
\newblock {\em ICLR}, 2018.

\bibitem{benaim2021structural}
Sagie Benaim, Ron Mokady, Amit Bermano, and L Wolf.
\newblock Structural analogy from a single image pair.
\newblock In {\em Computer Graphics Forum}, volume~40, pages 249--265. Wiley
  Online Library, 2021.

\bibitem{benaim2017one}
Sagie Benaim and Lior Wolf.
\newblock One-sided unsupervised domain mapping.
\newblock In {\em NeurIPS}, pages 752--762, 2019.

\bibitem{binkowski2018demystifying}
M Bi{\'n}kowski, DJ Sutherland, M Arbel, and A Gretton.
\newblock Demystifying mmd gans.
\newblock In {\em ICLR}, 2018.

\bibitem{bro2014principal}
Rasmus Bro and Age~K Smilde.
\newblock Principal component analysis.
\newblock {\em Analytical Methods}, 6(9):2812--2831, 2014.

\bibitem{brock2018large}
Andrew Brock, Jeff Donahue, and Karen Simonyan.
\newblock Large scale gan training for high fidelity natural image synthesis.
\newblock In {\em ICLR}, 2019.

\bibitem{chen2020reusing}
Runfa Chen, Wenbing Huang, Binghui Huang, Fuchun Sun, and Bin Fang.
\newblock Reusing discriminators for encoding towards unsupervised
  image-to-image translation.
\newblock In {\em CVPR}, 2020.

\bibitem{StarGAN2018}
Yunjey Choi, Minje Choi, Munyoung Kim, Jung-Woo Ha, Sunghun Kim, and Jaegul
  Choo.
\newblock Stargan: Unified generative adversarial networks for multi-domain
  image-to-image translation.
\newblock In {\em CVPR}, June 2018.

\bibitem{choi2020stargan}
Yunjey Choi, Youngjung Uh, Jaejun Yoo, and Jung-Woo Ha.
\newblock Stargan v2: Diverse image synthesis for multiple domains.
\newblock In {\em CVPR}, 2020.

\bibitem{cohen2019bidirectional}
Tomer Cohen and Lior Wolf.
\newblock Bidirectional one-shot unsupervised domain mapping.
\newblock In {\em ICCV}, pages 1784--1792, 2019.

\bibitem{denton2015deep}
Emily~L Denton, Soumith Chintala, Rob Fergus, et~al.
\newblock Deep generative image models using a laplacian pyramid of adversarial
  networks.
\newblock In {\em NeurIPS}, pages 1486--1494, 2015.

\bibitem{donahue2014decaf}
Jeff Donahue, Yangqing Jia, Oriol Vinyals, Judy Hoffman, Ning Zhang, Eric
  Tzeng, and Trevor Darrell.
\newblock Decaf: A deep convolutional activation feature for generic visual
  recognition.
\newblock In {\em ICML}, pages 647--655, 2014.

\bibitem{goetschalckx2019ganalyze}
Lore Goetschalckx, Alex Andonian, Aude Oliva, and Phillip Isola.
\newblock Ganalyze: Toward visual definitions of cognitive image properties.
\newblock In {\em ICCV}, pages 5744--5753, 2019.

\bibitem{gonzalez2018image}
Abel Gonzalez-Garcia, Joost van~de Weijer, and Yoshua Bengio.
\newblock Image-to-image translation for cross-domain disentanglement.
\newblock In {\em NeurIPS}, pages 1294--1305, 2018.

\bibitem{goodfellow2014generative}
Ian Goodfellow, Jean Pouget-Abadie, Mehdi Mirza, Bing Xu, David Warde-Farley,
  Sherjil Ozair, Aaron Courville, and Yoshua Bengio.
\newblock Generative adversarial nets.
\newblock In {\em NeurIPS}, pages 2672--2680, 2014.

\bibitem{gulrajani2017improved}
Ishaan Gulrajani, Faruk Ahmed, Martin Arjovsky, Vincent Dumoulin, and Aaron~C
  Courville.
\newblock Improved training of wasserstein gans.
\newblock In {\em NeurIPS}, pages 5767--5777, 2017.

\bibitem{heusel2017gans}
Martin Heusel, Hubert Ramsauer, Thomas Unterthiner, Bernhard Nessler, and Sepp
  Hochreiter.
\newblock Gans trained by a two time-scale update rule converge to a local nash
  equilibrium.
\newblock In {\em NeurIPS}, pages 6626--6637, 2017.

\bibitem{huang2018multimodal}
Xun Huang, Ming-Yu Liu, Serge Belongie, and Jan Kautz.
\newblock Multimodal unsupervised image-to-image translation.
\newblock In {\em ECCV}, pages 172--189, 2018.

\bibitem{pix2pix2017}
Phillip Isola, Jun-Yan Zhu, Tinghui Zhou, and Alexei~A Efros.
\newblock Image-to-image translation with conditional adversarial networks.
\newblock In {\em CVPR}, pages 1125--1134, 2017.

\bibitem{isola2016image}
Phillip Isola, Jun-Yan Zhu, Tinghui Zhou, and Alexei~A Efros.
\newblock Image-to-image translation with conditional adversarial networks.
\newblock In {\em CVPR}, 2017.

\bibitem{jahanian2019steerability}
Ali Jahanian, Lucy Chai, and Phillip Isola.
\newblock On the''steerability" of generative adversarial networks.
\newblock In {\em ICLR}, 2020.

\bibitem{jakab2018unsupervised}
Tomas Jakab, Ankush Gupta, Hakan Bilen, and Andrea Vedaldi.
\newblock Unsupervised learning of object landmarks through conditional image
  generation.
\newblock In {\em NeurIPS}, pages 4016--4027, 2018.

\bibitem{johnson2018image}
Justin Johnson, Agrim Gupta, and Li Fei-Fei.
\newblock Image generation from scene graphs.
\newblock In {\em CVPR}, pages 1219--1228, 2018.

\bibitem{karras2019style}
Tero Karras, Samuli Laine, and Timo Aila.
\newblock A style-based generator architecture for generative adversarial
  networks.
\newblock In {\em CVPR}, pages 4401--4410, 2019.

\bibitem{katzir2019cross}
Oren Katzir, Dani Lischinski, and Daniel Cohen-Or.
\newblock Cross-domain cascaded deep feature translation.
\newblock {\em arXiv}, pages arXiv--1906, 2019.

\bibitem{kawano2014automatic}
Yoshiyuki Kawano and Keiji Yanai.
\newblock Automatic expansion of a food image dataset leveraging existing
  categories with domain adaptation.
\newblock In {\em ECCV}, pages 3--17. Springer, 2014.

\bibitem{kim2019u}
Junho Kim, Minjae Kim, Hyeonwoo Kang, and Kwanghee Lee.
\newblock U-gat-it: unsupervised generative attentional networks with adaptive
  layer-instance normalization for image-to-image translation.
\newblock {\em arXiv preprint arXiv:1907.10830}, 2019.

\bibitem{kim2017learning}
Taeksoo Kim, Moonsu Cha, Hyunsoo Kim, Jungkwon Lee, and Jiwon Kim.
\newblock Learning to discover cross-domain relations with generative
  adversarial networks.
\newblock In {\em ICML}, 2017.

\bibitem{kirkpatrick2017overcoming}
James Kirkpatrick, Razvan Pascanu, Neil Rabinowitz, Joel Veness, Guillaume
  Desjardins, Andrei~A Rusu, Kieran Milan, John Quan, Tiago Ramalho, Agnieszka
  Grabska-Barwinska, et~al.
\newblock Overcoming catastrophic forgetting in neural networks.
\newblock {\em Proceedings of the national academy of sciences},
  114(13):3521--3526, 2017.

\bibitem{Lee2018drit}
Hsin-Ying Lee, Hung-Yu Tseng, Jia-Bin Huang, Maneesh~Kumar Singh, and
  Ming-Hsuan Yang.
\newblock Diverse image-to-image translation via disentangled representations.
\newblock In {\em ECCV}, 2018.

\bibitem{lee2020drit++}
Hsin-Ying Lee, Hung-Yu Tseng, Qi Mao, Jia-Bin Huang, Yu-Ding Lu, Maneesh Singh,
  and Ming-Hsuan Yang.
\newblock Drit++: Diverse image-to-image translation via disentangled
  representations.
\newblock {\em IJCV}, pages 1--16, 2020.

\bibitem{li2020few}
Yijun Li, Richard Zhang, Jingwan Lu, and Eli Shechtman.
\newblock Few-shot image generation with elastic weight consolidation.
\newblock In {\em NeurIPS}, 2020.

\bibitem{lin2020tuigan}
Jianxin Lin, Yingxue Pang, Yingce Xia, Zhibo Chen, and Jiebo Luo.
\newblock Tuigan: Learning versatile image-to-image translation with two
  unpaired images.
\newblock In {\em European Conference on Computer Vision}, pages 18--35.
  Springer, 2020.

\bibitem{liu2018unified}
Alexander~H Liu, Yen-Cheng Liu, Yu-Ying Yeh, and Yu-Chiang~Frank Wang.
\newblock A unified feature disentangler for multi-domain image translation and
  manipulation.
\newblock In {\em NeurIPS}, pages 2590--2599, 2018.

\bibitem{liu2017unsupervised}
Ming-Yu Liu, Thomas Breuel, and Jan Kautz.
\newblock Unsupervised image-to-image translation networks.
\newblock In {\em NeurIPS}, pages 700--708, 2017.

\bibitem{liu2019few}
Ming-Yu Liu, Xun Huang, Arun Mallya, Tero Karras, Timo Aila, Jaakko Lehtinen,
  and Jan Kautz.
\newblock Few-shot unsupervised image-to-image translation.
\newblock In {\em CVPR}, pages 10551--10560, 2019.

\bibitem{maaten2008visualizing}
Laurens van~der Maaten and Geoffrey Hinton.
\newblock Visualizing data using t-sne.
\newblock {\em Journal of machine learning research}, 9(Nov):2579--2605, 2008.

\bibitem{mao2017least}
Xudong Mao, Qing Li, Haoran Xie, Raymond~YK Lau, Zhen Wang, and Stephen
  Paul~Smolley.
\newblock Least squares generative adversarial networks.
\newblock In {\em ICCV}, pages 2794--2802, 2017.

\bibitem{mejjati2018unsupervised}
Youssef~Alami Mejjati, Christian Richardt, James Tompkin, Darren Cosker, and
  Kwang~In Kim.
\newblock Unsupervised attention-guided image-to-image translation.
\newblock In {\em NeurIPS}, pages 3693--3703, 2018.

\bibitem{mo2020freeze}
Sangwoo Mo, Minsu Cho, and Jinwoo Shin.
\newblock Freeze the discriminator: a simple baseline for fine-tuning gans.
\newblock In {\em CVPR AI for Content Creation Workshop}, 2020.

\bibitem{noguchi2019image}
Atsuhiro Noguchi and Tatsuya Harada.
\newblock Image generation from small datasets via batch statistics adaptation.
\newblock {\em ICCV}, 2019.

\bibitem{park2020contrastive}
Taesung Park, Alexei~A. Efros, Richard Zhang, and Jun-Yan Zhu.
\newblock Contrastive learning for conditional image synthesis.
\newblock In {\em ECCV}, 2020.

\bibitem{paszke2017automatic}
Adam Paszke, Sam Gross, Soumith Chintala, Gregory Chanan, Edward Yang, Zachary
  DeVito, Zeming Lin, Alban Desmaison, Luca Antiga, and Adam Lerer.
\newblock Automatic differentiation in pytorch.
\newblock 2017.

\bibitem{perarnau2016invertible}
Guim Perarnau, Joost Van De~Weijer, Bogdan Raducanu, and Jose~M {\'A}lvarez.
\newblock Invertible conditional gans for image editing.
\newblock In {\em NeurIPS}, 2016.

\bibitem{saito2020coco}
Kuniaki Saito, Kate Saenko, and Ming-Yu Liu.
\newblock Coco-funit: Few-shot unsupervised image translation with a content
  conditioned style encoder.
\newblock {\em arXiv preprint arXiv:2007.07431}, 2020.

\bibitem{shmelkov2018good}
Konstantin Shmelkov, Cordelia Schmid, and Karteek Alahari.
\newblock How good is my gan?
\newblock In {\em ECCV}, pages 213--229, 2018.

\bibitem{shocher2020semantic}
Assaf Shocher, Yossi Gandelsman, Inbar Mosseri, Michal Yarom, Michal Irani,
  William~T Freeman, and Tali Dekel.
\newblock Semantic pyramid for image generation.
\newblock In {\em CVPR}, pages 7457--7466, 2020.

\bibitem{shrivastava2017learning}
Ashish Shrivastava, Tomas Pfister, Oncel Tuzel, Joshua Susskind, Wenda Wang,
  and Russell Webb.
\newblock Learning from simulated and unsupervised images through adversarial
  training.
\newblock In {\em CVPR}, pages 2107--2116, 2017.

\bibitem{simonyan2013deep}
Karen Simonyan, Andrea Vedaldi, and Andrew Zisserman.
\newblock Deep inside convolutional networks: Visualising image classification
  models and saliency maps.
\newblock {\em arXiv preprint arXiv:1312.6034}, 2013.

\bibitem{van2015building}
Grant Van~Horn, Steve Branson, Ryan Farrell, Scott Haber, Jessie Barry, Panos
  Ipeirotis, Pietro Perona, and Serge Belongie.
\newblock Building a bird recognition app and large scale dataset with citizen
  scientists: The fine print in fine-grained dataset collection.
\newblock In {\em CVPR}, pages 595--604, 2015.

\bibitem{wang2019sdit}
Yaxing Wang, Abel Gonzalez-Garcia, Joost van~de Weijer, and Luis Herranz.
\newblock {SDIT}: Scalable and diverse cross-domain image translation.
\newblock In {\em ACM MM}, 2019.

\bibitem{wang2018transferring}
Yaxing Wang, Chenshen Wu, Luis Herranz, Joost van~de Weijer, Abel
  Gonzalez-Garcia, and Bogdan Raducanu.
\newblock Transferring gans: generating images from limited data.
\newblock In {\em ECCV}, pages 218--234, 2018.

\bibitem{wang2020deepi2i}
Yaxing Wang, Lu Yu, and Joost van~de Weijer.
\newblock Deepi2i: Enabling deep hierarchical image-to-image translation by
  transferring from gans.
\newblock {\em NeurIPS}, 2020.

\bibitem{wu2019transgaga}
Wayne Wu, Kaidi Cao, Cheng Li, Chen Qian, and Chen~Change Loy.
\newblock Transgaga: Geometry-aware unsupervised image-to-image translation.
\newblock In {\em CVPR}, pages 8012--8021, 2019.

\bibitem{yi2017dualgan}
Zili Yi, Hao Zhang, Ping~Tan Gong, et~al.
\newblock Dualgan: Unsupervised dual learning for image-to-image translation.
\newblock In {\em ICCV}, 2017.

\bibitem{yu2019multi}
Xiaoming Yu, Yuanqi Chen, Shan Liu, Thomas Li, and Ge Li.
\newblock Multi-mapping image-to-image translation via learning
  disentanglement.
\newblock In {\em NeurIPS}, pages 2990--2999, 2019.

\bibitem{zhao2020leveraging}
Miaoyun Zhao, Yulai Cong, and Lawrence Carin.
\newblock On leveraging pretrained gans for limited-data generation.
\newblock {\em ICML}, 2020.

\bibitem{zhu2020domain}
Jiapeng Zhu, Yujun Shen, Deli Zhao, and Bolei Zhou.
\newblock In-domain gan inversion for real image editing.
\newblock {\em ECCV}, 2020.

\bibitem{zhu2017unpaired}
Jun-Yan Zhu, Taesung Park, Phillip Isola, and Alexei~A Efros.
\newblock Unpaired image-to-image translation using cycle-consistent
  adversarial networks.
\newblock In {\em ICCV}, pages 2223--2232, 2017.

\bibitem{zhu2017toward}
Jun-Yan Zhu, Richard Zhang, Deepak Pathak, Trevor Darrell, Alexei~A Efros,
  Oliver Wang, and Eli Shechtman.
\newblock Toward multimodal image-to-image translation.
\newblock In {\em NeurIPS}, pages 465--476, 2017.

\end{thebibliography}
}

\end{document}